\documentclass[10pt,journal,compsoc]{IEEEtran}
\ifCLASSOPTIONcompsoc
  \usepackage[nocompress]{cite}
\else
  \usepackage{cite}
\fi
\ifCLASSINFOpdf
\else
\fi

\usepackage{epsfig}
\usepackage{graphicx}
\usepackage{amsmath}
\usepackage{amssymb}
\usepackage{enumitem}

\usepackage[pagebackref=false,breaklinks=true,letterpaper=true,colorlinks,bookmarks=false,urlcolor=red,citecolor=cyan]{hyperref}
\usepackage{gensymb,graphics,subfigure,inputenc}
\usepackage[linesnumbered,algo2e,boxed]{algorithm2e}
\graphicspath{{Figure/}}
\usepackage[figuresright]{rotating}
\usepackage{amsmath,amssymb,textcomp,subfigure,multirow,upgreek}
\usepackage{booktabs}
\usepackage{color,mdwlist}

\usepackage{ragged2e}

\usepackage{review}

\usepackage{graphicx,fontawesome}
\graphicspath{{Figures/}}

\usepackage{caption}
\usepackage{wrapfig}

\usepackage{enumitem}
\setitemize{noitemsep,topsep=0pt,parsep=0pt,partopsep=0pt}

\usepackage{multirow}   
\usepackage{booktabs}    
\usepackage{makecell}
\usepackage{graphicx} 
\usepackage{array} 
\usepackage{hyperref} 
\usepackage{tabularx} 
\usepackage{xspace} 
\usepackage{caption} 
\usepackage{lscape} 
\usepackage{amssymb} 

\newcolumntype{L}[1]{>{\raggedright\arraybackslash}p{#1}}
\newcolumntype{R}[1]{>{\raggedleft\arraybackslash}p{#1}}

\usepackage{arydshln}  

\begin{document}
%
\title{SAM2 for Image and Video Segmentation: A Comprehensive Survey}
\author{ Jiaxing Zhang \quad Hao Tang$^*$
	\IEEEcompsocitemizethanks{
	    \IEEEcompsocthanksitem  Jiaxing Zhang is with the School of Software engineering, Sichuan University, Chengdu 610065, China. E-mail: 2022141420175@stu.scu.edu.cn. \protect
        \IEEEcompsocthanksitem Hao Tang is with the School of Computer Science, Peking University, Beijing 100871, China. E-mail: haotang@pku.edu.cn \protect
        }
	\thanks{$^*$Corresponding author: Hao Tang.}
}

%
%

\markboth{IEEE Transactions on Pattern Analysis and Machine Intelligence}%
{Shell \MakeLowercase{\textit{et al.}}: Bare Demo of IEEEtran.cls for Computer Society Journals}
%



\IEEEtitleabstractindextext{%
\justify
\begin{abstract}

Despite significant advances in deep learning for image and video segmentation, existing models continue to face challenges in cross-domain adaptability and generalization. Image and video segmentation are fundamental tasks in computer vision with wide-ranging applications in healthcare, agriculture, industrial inspection, and autonomous driving.
With the advent of large-scale foundation models, SAM2—an improved version of SAM (Segment Anything Model)—has been optimized for segmentation tasks, demonstrating enhanced performance in complex scenarios. However, SAM2's adaptability and limitations in specific domains require further investigation.
This paper systematically analyzes the application of SAM2 in image and video segmentation and evaluates its performance in various fields. We begin by introducing the foundational concepts of image segmentation, categorizing foundation models, and exploring the technical characteristics of SAM and SAM2. Subsequently, we delve into SAM2's applications in static image and video segmentation, emphasizing its performance in specialized areas such as medical imaging and the challenges of cross-domain adaptability. As part of our research, we reviewed over 200 related papers to provide a comprehensive analysis of the topic.
Finally, the paper highlights the strengths and weaknesses of SAM2 in segmentation tasks, identifies the technical challenges it faces, and proposes future development directions. This review provides valuable insights and practical recommendations for optimizing and applying SAM2 in real-world scenarios.
\end{abstract}

}

\maketitle

\IEEEdisplaynontitleabstractindextext

%
\IEEEpeerreviewmaketitle


%
%
%
%

\section{Introduction} \label{introduction}


\IEEEPARstart{I}{mage} segmentation and video segmentation are fundamental tasks in computer vision, designed to partition images or videos into meaningful regions based on semantic or spatial features~\cite{szeliski2022computer}. These tasks have found applications in diverse fields, including healthcare~\cite{li2020color,zhou2019high}, agriculture~\cite{pantazi2019automated}, industrial inspection~\cite{lee2019steel}, autonomous driving~\cite{cordts2016cityscapes}, and satellite remote sensing~\cite{wang2019hybrid,zhang2019hierarchical}.
Image segmentation focuses on identifying and extracting target objects, boundaries, or textures from a single image, while video segmentation extends this process to the temporal dimension, aiming to segment consecutive video frames accurately while ensuring spatio-temporal consistency.
Recent advances in deep learning have led to significant breakthroughs in addressing these tasks, even in complex scenarios. However, most existing models are tailored to specific imaging modalities or tasks, which limits their ability to generalize effectively across diverse domains. Consequently, developing more general and adaptable segmentation models has emerged as a critical direction for advancing the field.

\begin{figure*}
    \centering
    \includegraphics[width=1\linewidth]{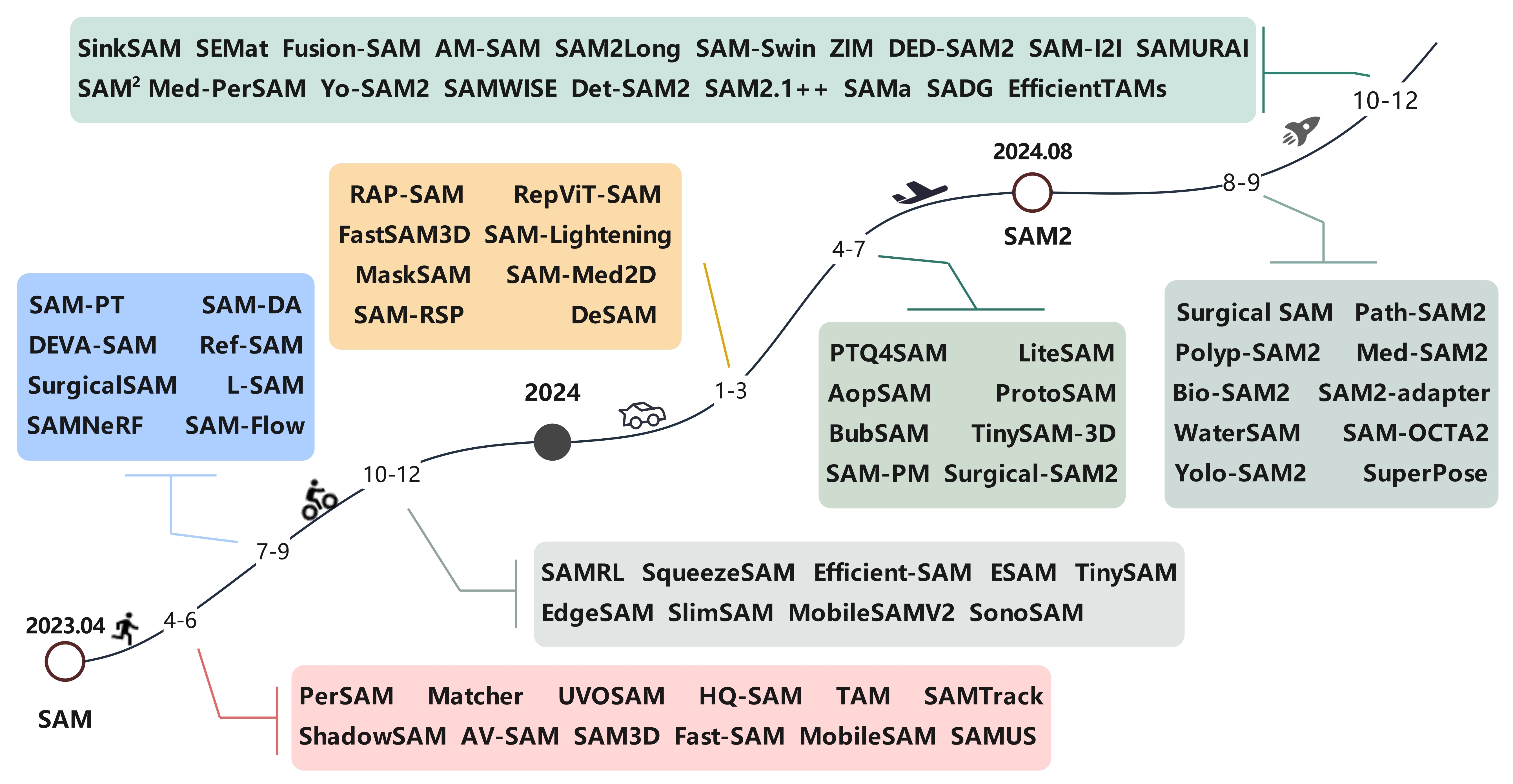}
    \caption{
    This image illustrates the evolution and categorization of the SAM (Segment Anything Model)/SAM2 and its derivatives. Different colors and positions are used to clearly represent each model along the timeline. In addition to compiling various SAM/SAM2 variants in the field of segmentation, including tasks such as shadow detection and classification, we place particular emphasis on the progression of segmentation tasks.
}
    \label{fig:Timeline}
\end{figure*}


The emergence of large-scale foundation models has revolutionized the research paradigm in artificial intelligence, demonstrating remarkable zero-shot and few-shot learning capabilities~\cite{wang2023large}. As a pioneering foundation model for image segmentation, SAM (Segment Anything Model)~\cite{SAM} has achieved notable success in natural image segmentation tasks. However, SAM faces several challenges when applied to image and video segmentation.
First, because SAM's pre-training primarily relies on natural images~\cite{compareSAMandSAM2,MedSAM,Sa-med2d-20m}, it struggles to adapt effectively to other domains, resulting in reduced accuracy. Second, SAM is predominantly trained on 2D images, which limits its performance when dealing with 3D medical images and other complex data types~\cite{MaSAM}. Lastly, SAM encounters difficulties in video segmentation tasks due to the temporal continuity and dynamic features inherent in video data~\cite{videosam}, which differ significantly from the requirements for static images.
To address these limitations, SAM2~\cite{SAM2}, an improved version of SAM, has been introduced. SAM2 is designed to better accommodate the specific needs of various segmentation tasks, offering more robust and accurate solutions for both image and video segmentation~\cite{compareSAMandSAM2}.

\begin{table}[t]
\caption{Comparative analysis of current surveys on universal image/video segmentation.}
\label{tab:comparison}
\centering
\tabcolsep=3pt
\resizebox{1.0\linewidth}{!}{
\begin{tabular}{ccccccc}
    \toprule
    {Survey} &  {\makecell[c]{SAM2 \\Focus}}& {\makecell[c]{SAM\\ Focus}} & {\makecell[c]{Video\\Segmentation}} &  {\makecell[c]{Medical \\ Scene}}& {\makecell[c]{Natural\\ Scene}}& {Year}\\
    \midrule
    \cite{zhang2023towards}& $\times$& $\checkmark$&$\times$& $\checkmark$& $\times$& 2023\\
    \cite{zhang2023comprehensive}& $\times$& $\checkmark$&$\times$& $\checkmark$& $\checkmark$& 2023\\
    \cite{Unleashing-SAM2Survey} & $\checkmark$& $\times$&$\times$& $\checkmark$& $\times$& 2024\\
    \cite{zhou2024sam2} & $\checkmark$& $\times$&$\checkmark$& $\times$& $\checkmark$& 2024\\
    \cite{sun2024efficientSAM} & $\times$& $\checkmark$&$\times$& $\checkmark$& $\times$& 2024\\
    \cite{zhang2024segment}& $\times$& $\checkmark$&$\checkmark$& $\times$& $\checkmark$& 2024\\
    Ours & $\checkmark$& $\checkmark$&$\checkmark$& $\checkmark$& $\checkmark$& 2024\\
    \bottomrule
\end{tabular}}
\vspace{-0.4cm}
\end{table}


To gain a deeper understanding of SAM2's role in image and video segmentation and provide a comprehensive perspective, we conducted a systematic review of relevant research~\cite{Unleashing-SAM2Survey,zhou2024sam2,zhang2023towards,zhang2024segment,zhang2023comprehensive}. Although some surveys summarize segmentation methods based on SAM or SAM2, these reviews often focus on specific domains or problems, overlooking the broader applications of SAM2 in both image and video segmentation (see Table \ref{tab:comparison}). This review is the first to comprehensively evaluate SAM2’s performance, highlighting its effectiveness in segmentation tasks, while also examining its adaptability and limitations across different domains.


This study focuses on analyzing SAM2’s performance in image and video segmentation tasks across various fields. First, we provide a comprehensive overview of image segmentation, foundational model concepts and classifications, and the technical characteristics of SAM and SAM2. We also discuss the efforts to extend SAM/SAM2 into other domains. Next, we summarize recent research advancements and evaluate SAM2’s segmentation performance in two primary areas: video and static images. In analyzing its performance on natural images, we place particular emphasis on its application in the specialized field of medical imaging, as studies in other specialized fields remain limited. Finally, we summarize SAM2’s characteristics in image and video segmentation, discuss current technical challenges, and explore future development directions.

The main objective of this study is to evaluate SAM2’s performance in image and video segmentation tasks. Section \ref{preliminaries} introduces the fundamental concepts of segmentation, covering the basics of image segmentation, the classification of foundational models, and a detailed comparison of SAM and SAM2, emphasizing their strengths and differences across tasks. Section \ref{image} reviews the latest research and applications of SAM2 in image segmentation. We examine state-of-the-art networks, summarize SAM- and SAM2-based methods, collect datasets for natural and medical images, and discuss commonly used evaluation metrics to establish a theoretical foundation for performance assessment.
In Section \ref{Video}, we shift focus to video segmentation tasks and evaluate SAM2’s performance in dynamic scenes. We classify recent video segmentation networks, compile relevant video datasets, and introduce evaluation metrics in this domain to provide a comprehensive analysis of SAM2’s capabilities. Finally, Section \ref{discussion} summarizes SAM2’s characteristics in image and video segmentation, identifies the technical challenges it faces, and offers insights into future development directions.
Through this study, our objective is to provide valuable information and actionable recommendations for the application and further optimization of SAM2 in real-world scenarios.

\section{Preliminaries}\label{preliminaries}

In this section, we provide a concise introduction to problem formulation, key topics, and concepts, aiming to enhance comprehension of our work. 

\subsection{Image Segmentation and Video Segmentation}


\textbf{Image segmentation} has been a foundational problem in computer vision since the early days of the field. It can be defined as the task of classifying pixels with semantic labels (semantic segmentation), partitioning individual objects (instance segmentation), or addressing both tasks simultaneously (panoptic segmentation).
Semantic segmentation\cite{chen2014semantic,chen2017rethinking,chen2018encoder} assigns each pixel to a predefined semantic class label. Instance segmentation\cite{hafiz2020survey,liu2018path,bolya2019yolact} goes a step further by distinguishing between instances of the same class. Panoptic segmentation, introduced by~\cite{kirillov2019panoptic}, integrates semantic and instance segmentation to provide a comprehensive understanding of scenes.
As the demand for more precise and user-friendly segmentation techniques has increased, \textbf{interactive segmentation} has emerged as a key approach in this domain. Interactive segmentation enables users to actively participate in the segmentation process by providing input and feedback, such as marking regions of interest or correcting errors. This interactive involvement aims to improve both the accuracy and efficiency of segmentation results. Unlike traditional semantic or instance segmentation methods, interactive segmentation empowers users to refine outputs dynamically, making it particularly effective in complex and ambiguous scenes.



\textbf{Video segmentation} is a crucial task in computer vision, aimed at classifying or segmenting each pixel in every frame of a video sequence into distinct objects and backgrounds~\cite{DAVIS16}. It can be broadly categorized into two types: Video Object Segmentation (VOS) and Video Semantic Segmentation (VSS)~\cite{zhou2022survey}. VOS focuses on isolating specific objects in a video without requiring detailed semantic labels, whereas VSS assigns each pixel to predefined semantic categories (e.g., ``person'' or ``car'') while ensuring label consistency across frames.
In recent years, \textbf{zero-shot} and \textbf{few-shot} learning approaches have attracted significant attention in video segmentation. Zero-shot learning performs segmentation tasks without requiring any training samples, whereas few-shot learning enables effective model training using only a small number of labeled samples.

While image segmentation is well-suited for static scenarios with lower computational complexity, video segmentation involves processing multiple frames, making it computationally intensive. It is specifically designed for dynamic scenarios that require higher performance and advanced techniques to handle the added complexity.

\subsection{Foundation Models}

\begin{figure}
    \centering
    \includegraphics[width=1\linewidth]{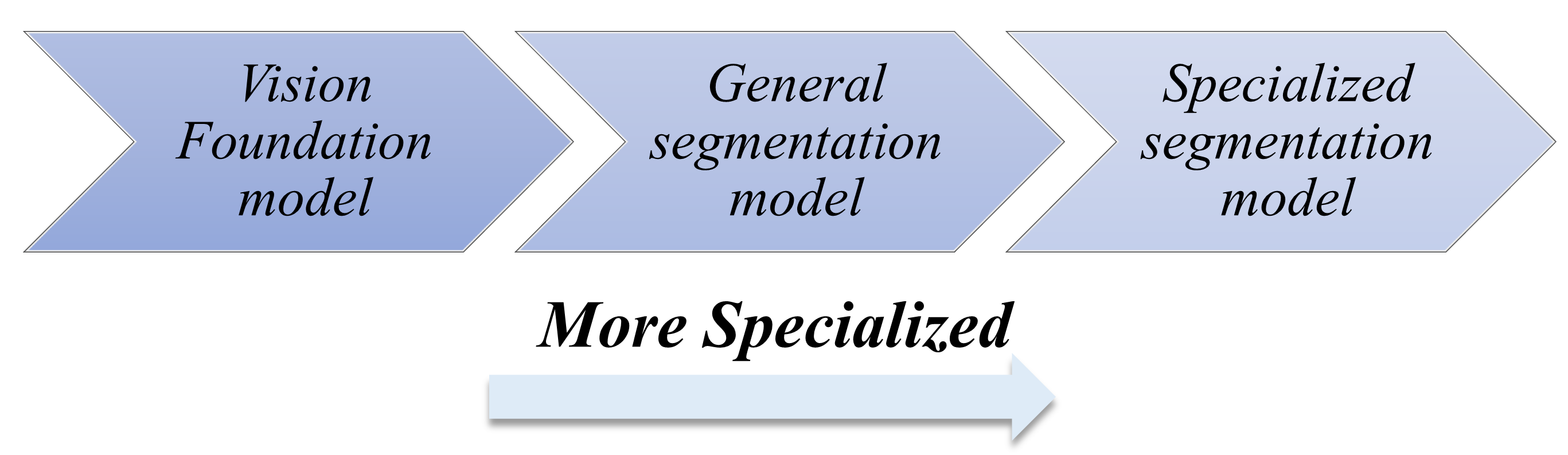}
    \caption{this image illustrates the hierarchical structure of visual models, progressing from foundational visual models to general segmentation models, and then to specialized segmentation models, reflecting an increasing level of specialization to meet the needs of more specific visual tasks.}
    \label{fig:foundationModel}
\end{figure}

Foundation models (e.g., \cite{bommasani2021opportunities}) refer to machine learning models that perform well on specific tasks and are pre-trained on large-scale datasets. Their design aims to enhance the model's generalizability and transferability, enabling it to perform well across various downstream tasks. As shown in Figure \ref{fig:foundationModel}, the trend towards increasingly refined models is presented. The characteristics of foundation models can be summarized with three keywords: large-scale pre-training, generalizability, and transferability.

\subsubsection{Vision Foundation Models}

The Visual Foundation Model \cite{yuan2021florence} is an advanced deep learning architecture designed to handle various visual tasks through a unified framework. These models are typically pre-trained on large-scale datasets to learn rich visual features, thereby enhancing their performance across multiple downstream tasks. The design of visual foundation models aims to achieve efficient feature extraction and multimodal understanding, enabling their application in fields such as image classification, object detection, image segmentation, and image generation. A notable example is CLIP \cite{radford2021learning}, which uses contrastive learning to map images and text into the same feature space, allowing it to understand and generate textual descriptions related to visual content. Additionally, the Vision Transformer (ViT) \cite{alexey2020image,han2022survey} effectively processes image data using self-attention mechanisms, adapting to various visual tasks. The emergence of these models has accelerated the rapid development of computer vision, providing strong support for practical applications.

\subsubsection{General Segmentation Models}

A general segmentation model is designed to handle a variety of objects and tasks within image segmentation, capable of addressing various types of image segmentation tasks through a unified framework. The most typical example is the Segment Anything Model \cite{SAM}, which is the first promptable general image segmentation model. SAM exemplifies the capabilities of foundation models, showcasing outstanding generalizability through pre-training on large-scale datasets and the ability to adapt to different domains via transfer learning. Consequently, this model shows extensive potential in applications such as medical imaging, autonomous driving, and robotic vision, providing fast and accurate segmentation results that enhance task efficiency and accuracy. Leveraging an efficient self-attention mechanism, SAM quickly responds to various scenarios, meeting modern applications' demands for real-time processing.

Building on this, SAM2~\cite{SAM2} further enhances the model's performance and flexibility. It introduces more complex feature extraction techniques and improved multimodal learning capabilities, allowing it to excel in handling complex backgrounds and diverse objects. Through optimized training strategies, SAM2 achieves high-quality segmentation with less annotated data, demonstrating greater adaptability and ease of integration with other systems. Additionally, SAM2 supports more detailed segmentation and higher resolutions, further improving precision and reliability in specific tasks. These enhancements make SAM2 more efficient in practical applications, capable of meeting the growing demands for real-time processing and high precision.

\subsubsection{Comparison of Specialized Models and Foundation Models}

From a definitional standpoint, specialized models possess in-depth expertise on specific topics, while foundation models encompass broader domain knowledge, whether within a single field or across multiple disciplines. Taking medical image segmentation as an example, \textbf{specialized foundational segmentation models} are often optimized for specific organs or tasks, thus providing higher accuracy and interpretability in relevant segmentation tasks. In contrast, \textbf{general segmentation models} serve as a multitasking and multimodal platform that can process multimodal images of various organs and diseases. Our goal is to leverage general segmentation models to achieve convenient segmentation functionality while reaching or even surpassing the precision levels of specialized segmentation models.

\subsection{Evolution of Image Segmentation}

\subsubsection{SAM: Segment Anything}

\begin{figure}
    \centering
    \includegraphics[width=1\linewidth]{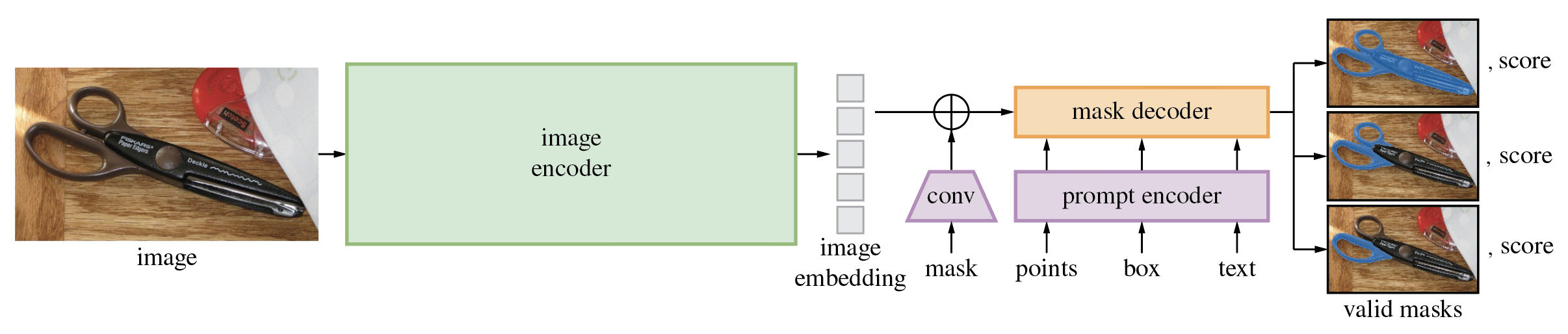}
    \caption{Segment anything~\cite{SAM}}
    \label{sam}
\end{figure}

As shown in Figure \ref{sam} , Segmen anything \cite{SAM}, as the first foundational model proposed for promptable image segmentation, consists primarily of three main components: the Image Encoder, Mask Decoder, and Prompt Encoder.

\textbf{Image encoder} use a MAE \cite{he2022masked} pre-trained Vision Transformer (ViT) \cite{han2022survey} minimally adapted to process high resolution inputs.

\textbf{Prompt encoder} is crafted to process diverse user inputs—such as points, boxes, or text—by incorporating positional encodings\cite{tancik2020fourier} to guide the segmentation process. It encodes these prompts into a feature space aligned with the image features generated by the image encoder, enabling seamless integration and effective guidance during segmentation.

\textbf{Mask decoder} utilizes a modified Transformer decoder block \cite{vaswani2017attention}, followed by a dynamic mask prediction head, to generate segmentation masks by effectively combining image embeddings with prompt embeddings.

\subsubsection{SAM2: Segment Anything in Images and Videos}

\begin{figure}
    \centering
    \includegraphics[width=1\linewidth]{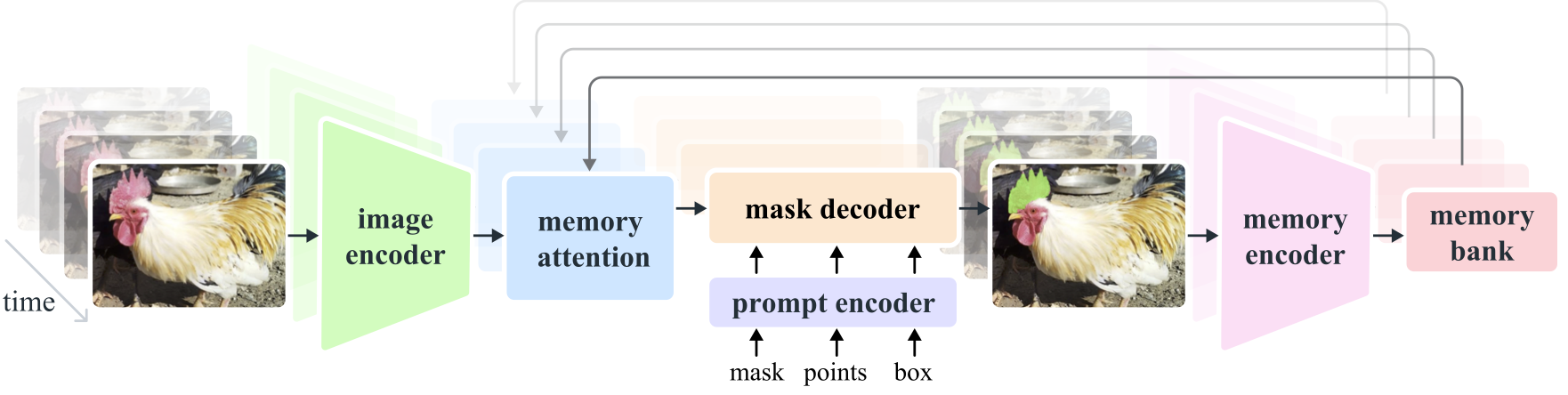}
    \caption{SAM2~\cite{SAM2}}
    \label{sam2}
\end{figure}

As shown in Figure \ref{sam2}, Segment Anything Model 2 (SAM2) \cite{SAM2} is an advanced visual segmentation model that builds on its predecessor, SAM \cite{SAM}, by incorporating a transformer-based architecture combined with a streaming memory component. This enhancement allows SAM2 to support real-time video segmentation and object tracking, addressing the dynamic challenges posed by moving scenes. 
SAM2's architecture is optimized for real-time video segmentation and object tracking, featuring several key components. The \textbf{Hierarchical Image Encoder} performs initial feature extraction, generating unconditioned tokens for each frame, and running once per interaction to provide frame representations. The \textbf{Memory Attention Module} leverages temporal context by conditioning current frame features on those from past frames, along with prior predictions and any new prompts, using efficient self- and cross-attention mechanisms \cite{dao2023flashattention}.

To handle user input, the \textbf{Prompt Encoder}—identical to SAM's—interprets prompts like clicks (positive or negative), bounding boxes, or masks to specify the object's extent within a frame. The \textbf{Mask Decoder} then generates segmentation masks, maintaining the SAM's approach for continuity. The \textbf{Memory Encoder} refines memory features by downsampling output masks and integrating them with unconditioned frame embeddings through lightweight convolutional layers. Lastly, the \textbf{Memory Bank} stores past predictions, enhancing accuracy and reducing the need for user input. Together, these components boost SAM2's segmentation accuracy and efficiency, reducing interactions, and significantly speeding up image segmentation tasks. SAM-2 has demonstrated impressive zero-shot performance in various tasks that involve natural image and video segmentation.

\subsubsection{Comparison between SAM and SAM2}

There are significant differences between SAM and SAM2 in terms of applicable scope, architecture, and real-world application scenarios \cite{geetha2024sam,compareSAMandSAM2}. In terms of \textbf{architecture}, SAM consists of a basic combination of an image encoder, mask decoder, and prompt encoder, while SAM2 builds on this foundation with the addition of a streaming memory component, which includes a hierarchical image encoder, memory attention module, and memory bank. These modules enable SAM2 to maintain consistency and accuracy across multiple frames. Regarding \textbf{prompting methods}, both SAM and SAM2's prompt encoders support inputs such as points, boxes, and text, but SAM2 leverages memory from previous frames in multi-frame environments to reduce user input. For \textbf{application scenarios}, SAM is commonly used for single-frame segmentation tasks, such as object recognition, while SAM2 supports object tracking across video sequences, making it particularly suitable for scenarios requiring temporal continuity, such as video editing and dynamic object detection in autonomous driving. In terms of \textbf{applicable scope}, SAM primarily targets static image segmentation tasks, such as object segmentation in single-frame images, useful in areas like medical image analysis or satellite image segmentation. SAM2, on the other hand, extends further into video segmentation and object tracking, focusing on real-time processing in dynamic scenes, making it well suited for continuous-frame tasks such as autonomous driving and video surveillance.

 \section{SAM2 for Image} \label{image}

Image segmentation divides the input image or video into multiple regions to distinguish different objects or structures. Accurate segmentation results help extract the regions or objects of interest and provide a foundation for subsequent analysis and processing, which is crucial for advanced computer vision tasks such as semantic understanding.

This subsection provides a comprehensive overview of image segmentation. We have compiled the current state-of-the-art models, particularly those based on SAM and SAM2. In addition, we summarize commonly used datasets and evaluation metrics for assessing image segmentation methods. 

\subsection{Image Segmentation Networks}

Table~\ref{tab:image-networks} presents a convenient reference for understanding the future of image segmentation in the field of deep learning by presenting segmentation methods from recent years.

Initially, early image segmentation methods relied on traditional machine learning techniques, typically using hand-crafted features for segmentation. With the advent of convolutional neural networks, image segmentation entered a new phase, where researchers began developing specialized segmentation models that achieved state-of-the-art performance on specific datasets. In recent years, the general-purpose segmentation model SAM has demonstrated significant superiority, providing more accurate and robust segmentation results in various scenarios. To further enhance segmentation precision and efficiency, SAM2 was introduced, offering faster processing times and more accurate segmentation, particularly in complex scenes and resource-constrained environments. SAM2 excels at handling large-scale datasets and real-time applications, driving significant advancements in the field of image segmentation. In the following subsections, we provide a detailed description of the characteristics and applications of these methods.

\begin{table*}
\renewcommand{\arraystretch}{1.2}

\centering
\caption{Recent advancements in image-segmentation networks, categorizing them by their foundational architecture and highlighting their unique features and targeted applications, with a particular emphasis on the evolution and capabilities of SAM and SAM2-based methodologies.}
\label{tab:image-networks}
\begin{tabular}{ccccccl}
\toprule
  Classification&Method&Year & Pub.& Backbone &Feature&Tasks\\
 \midrule
    \multirow{12}{*}{SAM-based}
   &MedSAM~\cite{MedSAM}&2023& Nature&  SAM backbone&Fine-tuned&universal MIS\\ 
   &MobileSAM~\cite{mobilesam}&2023& RS&  SAM backbone&Lightweight, Efficient&sVOS\\
&SAM-Aug~\cite{dai2023samaug}&2023& Arxiv&  SAM backbone&Point Augmentation&iVOS\\
   &SAM-Med2D~\cite{Sa-med2d-20m}&2023& Arxiv&  SAM backbone& Large-scale& universal MIS\\ 
 & PerSAM~\cite{PerSAM}& 2023& Arxiv& SAM backbone& personalization& Personalized imageSeg\\
 & SAM-Adapter~\cite{Sam-adapter}&2023& ICCV& SAM backbone& Adaptability& downstream tasks\\
&IMIS-Net~\cite{IMIS}&2024& BIBM& SAM backbone &pre-trained&interactive MIS\\
&Fusion-SAM~\cite{fusionsam}&2024& Arxiv& SAM backbone & multi-modal&autonomous driving\\
&Point-SAM~\cite{liu2024pointsam}&2024& Arxiv& SAM backbone &Point-supervised& RSIS\\
&SAM-MPA~\cite{SAM-mpa}&2024& Arxiv&  SAM backbone&K-centroid clustering&Few-shot\\
&CAT-SAM~\cite{cat-sam}&2024& ECCV&  SAM backbone&Efficient Adaptation&Zero-shot\\
&EVF-SAM~\cite{zhang2024evf}&2024& Arxiv&  SAM backbone&multi-modal &universal IS\\
\specialrule{0em}{1pt}{1pt}
\hdashline
\specialrule{0em}{1pt}{1pt}
\multirow{7}{*}{SAM2-based}&MedSAM2~\cite{medsam2}&2024& Arxiv&  SAM2 backbone&OnePrompt Seg&universal 2D and 3D MIS\\
 & Bio-SAM2~\cite{biosam2}& 2024& Arxiv& SAM2 backbone& Enhanced Biomedical Model& universal MIS\\
 & SAM-OCTA 2~\cite{sam2OCTA}& 2024& Arxiv& SAM2 backbone& Fine-tuned& Zero-shot\\
  & DED-SAM~\cite{ded-sam}& 2024& IEEE J-stars& SAM2 backbone& Change detection& RSIS\\
 & SAM2-Adapter~\cite{sam2Adapter}& 2024& Arxiv& SAM2 backbone&  Broad adaptability&Downstream tasks\\
 & PATH-SAM2~\cite{pathsam2}& 2024& Arxiv& SAM2 backbone& Fine-tuned& digital pathology\\

  \specialrule{0em}{1pt}{1pt}
\hdashline
\specialrule{0em}{1pt}{1pt}

 \multirow{7}{*}{Others}& nn-Unet~\cite{nnUnet}& 2021& Nature& Unet& Automation, Generality& task-tailored\\
 & TransUnet~\cite{transunet}& 2021& Arxiv& Transformer, Unet& Hybrid architecture&task-tailored\\
&UniverSeg~\cite{universeg}& 2023& ICCV& FCN& Zero-shot generalization& universal MIS\\
&FCFI~\cite{FCFI}& 2023& CVPR& ResNet-101 & Feedback-Driven& interactive imageSeg\\
&iMOS~\cite{iMOS}& 2024& ISBI& XMem~\cite{Xmem}&Minimal annotation& Few-shot\\
 & MedSegDiff~\cite{medsegdiff}& 2024& MIDL& diffusion model& Conditional diffusion&task-tailored\\
&OnePrompt~\cite{onePrompt}& 2024& CVPR& CNN/ ViT& Single prompt& interactive MIS, Zero-shot\\
\bottomrule
\end{tabular}
\end{table*}

\subsubsection{Specialized SOTA Models}

Specialized \textit{state-of-the-art} models are designed with specific tasks in mind. These models are typically optimized for particular datasets or applications, achieving high performance in controlled environments. By focusing on a narrow range of problems, specialized models can outperform general-purpose approaches in terms of accuracy and efficiency for specific segmentation tasks. 
\begin{itemize}
    \item nn-Unet~\cite{nnUnet} is an automated medical image segmentation method based on U-Net, capable of automatically optimizing the network structure and adapting to various medical image tasks.
    \item MedSegDiff~\cite{medsegdiff}, based on Diffusion Probabilistic Models (DPM), enhances medical image segmentation performance with Dynamic Conditional Encoding and Feature Frequency Parser, widely applied to different medical tasks. 
    \item TransUnet~\cite{transunet} combines Transformers and U-Net, leveraging Transformers to capture global context, while U-Net enhances local details, improving medical image segmentation accuracy.
    \item iMOS~\cite{iMOS} is a medical image segmentation model based on Diffusion Models, designed for dynamic object segmentation, achieving efficient tracking and segmentation with minimal annotations.
    \item UniverSeg~\cite{universeg} performs precise medical image segmentation without additional training, using the CrossBlock mechanism with a query image and a sample set, suitable for unseen tasks.
    \item FCFI~\cite{FCFI} focuses on and collaboratively updates feedback and deep features, maximizing the use of feedback information in interactive segmentation to improve segmentation accuracy and efficiency.
    \item OnePrompt~\cite{onePrompt} combines single annotation and interactive segmentation methods, handling unseen tasks without training, demonstrating zero-shot capabilities in medical image segmentation.
\end{itemize}
These task-specific models are commonly used as baselines to evaluate the performance of general-purpose models. Since they are fine-tuned for specific tasks and provide high-accuracy segmentation results, they are often used as a standard to assess the effectiveness of new models. In comparative evaluations, general-purpose models must demonstrate performance comparable to or surpassing that of task-specific models to prove their effectiveness and robustness across diverse applications.

\subsubsection{SAM-based Models}

SAM-based models (Segment Anything Model) are general-purpose segmentation frameworks that provide a flexible and robust solution across a wide range of segmentation challenges. Unlike task-specific models, SAM is designed to handle diverse datasets, offering improved adaptability and accuracy in real-world applications. These models have set new benchmarks in the segmentation field by simplifying the task of segmenting complex scenes. 

\begin{itemize}
\item MedSAM~\cite{MedSAM} adapts the Segment Anything Model (SAM) for medical image segmentation, using domain-specific prompts to enhance segmentation performance on medical tasks and improve generalization in medical images.
\item SAM-Med2D~\cite{Sa-med2d-20m} is a large-scale 2D medical image segmentation dataset that integrates diverse medical images and the corresponding masks, which helps the development of specialized segmentation models for medical applications.
\item PerSAM~\cite{PerSAM} is a personalization method for SAM that enables task-specific segmentation with a single reference mask and image, adapting SAM without retraining and achieving personalized segmentation.
\item SAM-Adapter~\cite{Sam-adapter} enhances SAM's performance on challenging downstream tasks by incorporating task-specific knowledge using simple adapters, significantly improving segmentation results for specialized tasks.
\item SAM-MPA~\cite{SAM-mpa} is an extension of SAM for multi-modal medical image segmentation, improving performance by combining modality-specific information to adapt SAM's segmentation capabilities across different imaging techniques.
\item IMIS-Net~\cite{IMIS} designed for interactive medical image segmentation, utilizing dense mask generation through interactive inputs such as clicks, bounding boxes, and text prompts. It achieves superior accuracy and scalability in medical image segmentation tasks.
\item EVF-SAM~\cite{zhang2024evf} combines visual and text prompts, enhancing SAM's performance in referring segmentation tasks through early visual-language fusion, reducing parameters, and achieving higher performance.
\end{itemize}

\subsubsection{SAM2-based Models}

SAM2-based models represent the next evolution of the SAM framework, offering enhanced performance in terms of both speed and segmentation precision. Using advanced techniques and optimized architectures, SAM2 is capable of achieving accurate segmentation in complex and resource-constrained environments. This model is particularly well-suited for large-scale datasets and real-time applications, further advancing the capabilities of image segmentation.

\begin{itemize}
\item MedSAM2~\cite{medsam2}, the evolved version of MedSAM, designed to further enhance the segmentation of medical images. It offers improved accuracy and efficiency for medical image processing tasks, incorporating advanced techniques to handle complex anatomical structures more effectively. 
\item Bio-SAM2~\cite{biosam2}, focuses on the segmentation of biological images, incorporating domain-specific knowledge to achieve precise and adaptive segmentation for various biological imaging modalities.
\item SAM2-Adapter~\cite{sam2Adapter} is an adapter for SAM2 that integrates domain-specific knowledge and visual prompts, improving SAM2's performance in challenging segmentation tasks in medical imaging.
\item SAM-OCTA2~\cite{sam2OCTA} is tailored for Optical Coherence Tomography Angiography (OCTA) images, SAM-OCTA 2 enhances the segmentation of vascular structures with specialized techniques for OCTA data.
\item Path-SAM2~\cite{pathsam2} designed for pathology images, utilizing advanced adaptations of SAM to accurately segment histopathological structures for medical diagnosis.
\end{itemize}

In addition to these variations, existing works have systematically summarized the performance of SAM2 in medical imaging. Sengupta et al.~\cite{compareSAMandSAM2} compares the advancements of SAM2 relative to SAM in medical image segmentation, Dong et al.~\cite{dong2024segment} explores the performance of SAM2 on 3D images, Shen et al.~\cite{shen2024interactive} investigates SAM2's performance in interactive segmentation, He et al.~\cite{he2024short} evaluates SAM2's performance in 3D CT image segmentation, with results being less than ideal, Xiong et al.~\cite{xiong2024sam2} attempts to use SAM2 as an encoder for UNet, demonstrating its strong adaptability in image segmentation,  Ma et al.~\cite{ma2024segment} quickly adapts SAM2 to the medical domain through fine-tuning, Yamagishi et al.~\cite{yamagishi2024zero} evaluates SAM2's zero-shot performance on the abdominal organ CT scan dataset, Yildiz et al.~\cite{yildiz2024sam} adapts SAM2 for annotating 3D medical images and implements it through the 3D Slicer extension, supporting prompt-based annotation generation and propagation across volumes, Zhao et al.~\cite{zhao2024inspiring} focuses on the context dependence of SAM2. These evaluations have pushed the expansion of SAM2 across different domains.

\subsection{Datasets}

Next, we will specifically discuss widely used datasets for model training and evaluation, focusing on natural and medical scenes. In terms of natural scenes, there is already a wealth of research and datasets, so we will not go into further detail. However, in medical scenes, different datasets have varying requirements for models, so we will focus on this aspect.

\subsubsection{Natural datasets}

Natural scene datasets are widely used in computer vision research, supporting tasks such as object detection, segmentation, and scene understanding. These datasets covers various environments, from urban to natural settings, offering diverse data for model training. Specific dataset information is provided in Table~\ref{tab:natural-image-datasets}, and we will now provide a detailed introduction below.

\begin{table*}
\setlength\tabcolsep{1pt}
  \caption{Natural image datasets, detailing categories, sizes, annotations, and publication sources, aiding dataset selection for computer vision tasks, especially image segmentation.}
  \label{tab:natural-image-datasets}
  \centering
\begin{tabular}{c c c c c c c }
    \toprule
    {\makecell[c]{Category}}& {\makecell[c]{Dataset}}&  {\makecell[c]{Description}}& \makecell[c]{\# masks  type}& \makecell[c]{\# images  sampled}& \makecell[c]{\# masks  sampled} &{\makecell[c]{Pub.}}
\\
    \midrule
    
\multirow{3}{*}{UnderWater}& TrashCan~\cite{trashcan}& Underwater trash& Instance& 5936& 9540& Arxiv2020\\
& NDD20~\cite{ndd20}&  Above/underwater dolphin& Instance& 4402& 6100& Arxiv2020\\
& CoralVOS~\cite{CoralVOS}& Dense coral& Panoptic& 60456& 60456& Arxiv2023\\

\specialrule{0em}{1pt}{1pt}
\hdashline
\specialrule{0em}{1pt}{1pt}

\multirow{3}{*}{Plants}& PPDLS~\cite{PPDLS}& Tobacco leaves& Instance& 182& 2347& PRL2016\\

& MSU-PID~\cite{MSU-PID}& Arabidopsis growth& Instance& 900& 900& MVA2016\\

& KOMATSUNA~\cite{KOMATSUNA}&  Komatsuna leaves& Instance& 576& 576& ICCV2017\\

\specialrule{0em}{1pt}{1pt}
\hdashline
\specialrule{0em}{1pt}{1pt}

\multirow{10}{*}{Scene}

& SUN~\cite{SUN}& everyday scenes& Semantic& 130519& -&CVPR2010\\
& COCO~\cite{COCO}& Object detection, segmentation& Instance& 330K& 2.5M&ECCV2014\\

& Cityscapes~\cite{cityscapes}& Urban street scene& Panoptic& 293& 9973&CVPR2016\\
& Places~\cite{Places}& Scene object, diverse settings& Semantic& 10624928& 7076580&TPAMI2017\\

& ADE20K~\cite{ADE20K}& range of scenes and objects& Instance & 302& 10128& IJCV2019\\

& LVIS~\cite{LVIS}& Long-tailed object& Instance& 945& 9642&CVPR2019\\

& STREETS~\cite{streets}& Street-level video& Instance& 819& 9854&NIPS2019\\

& iShape~\cite{ishape}& Shape and structure, indoor& Instance& 754& 9742&Arxiv2021\\

& NDISPark~\cite{NDISPark1,NDISPark2}& parking lot footage& Instance& 111& 2577&IJCV2021\\

& IBD~\cite{IBD}& individual cell structures& Instance& 467& 11953&IEEE IGRS 2022\\

\specialrule{0em}{1pt}{1pt}
\hdashline
\specialrule{0em}{1pt}{1pt}

\multirow{3}{*}{Egocentric}
& GTEA~\cite{gtea}& human-object interactions& Instance& 652& 12.8& CVPR2015\\
& VISOR~\cite{VISOR}& complex real-world scenes& Instance& 1864& 10141&NIPS2022\\
& EgoHOS~\cite{EgoHos}& Ego-centric hand-object& Instance& 2970& 9961& ECCV2022\\

\specialrule{0em}{1pt}{1pt}
\hdashline
\specialrule{0em}{1pt}{1pt}

\multirow{3}{*}{Paintings}& DRAM~\cite{DRAM}& Art paintings& Semantic& 718& 1179& CGF2022\\

& SegCLP~\cite{SegCLP}& Chinese paintings& Semantic& 709& 709&DSP2024\\

& CLP~\cite{CLP}& Chinese landscape paintings& Semantic& 2207& -&NCA2024\\

\specialrule{0em}{1pt}{1pt}
\hdashline
\specialrule{0em}{1pt}{1pt}

\multirow{2}{*}{Zero-shot}
& SA-1B~\cite{SAM}& dense masks& Semantic& 11M& 1.1B& ICCV2023\\
& MicroMat-3K~\cite{zim}& Micro-level matte labels& Semantic& 3000& - & Arxiv2024\\

\specialrule{0em}{1pt}{1pt}
\hdashline
\specialrule{0em}{1pt}{1pt}

\multirow{8}{*}{Others}& BBBC038v1~\cite{BBBC038v1}& Biological nuclei& Instance& 227& 10506& Natures2019\\

& WoodScape~\cite{woodscape}& Surround fisheye& Instance& 107& 10266& NIPS2019\\
& DOORS~\cite{DOORS}& doorways, architectural elements& Instance& 10000& 10000& Zenodo2022\\

& TimberSeg~\cite{TimberSeg}& Operator-view timber logs& Instance& 220& 2487& IROS2022\\

& Hypersim~\cite{hypersim}& Photorealistic indoor masks& Instance& 338& 9445& ICCV2021\\

& Plittersdorf~\cite{Plittersdorf}&Wildlife traps& Instance& 187& 546& Sensors2022\\
 & ZeroWaste-f~\cite{zerowaste}& Recycling waste& Instance& 2947& 6155& CVPR2022\\
 & OVIS~\cite{OVIS}& Occlusion in videos& Instance& 2044& 10011& ICCV2023\\

    \bottomrule
\end{tabular}
\vspace{-3mm}
\end{table*}

\begin{itemize}
\item \textbf{UnderWater} datasets focus on underwater segmentation, addressing challenges such as lighting, turbidity, and distortions. For example, TrashCan~\cite{trashcan} collects data on underwater trash, while CoralVOS~\cite{CoralVOS} focuses on segmenting dense coral images. NDD20~\cite{ndd20} covers both underwater and above-water dolphin imagery, aiming to segment marine life across environments.
\item \textbf{Plants} datasets are used for segmentation and classification in botanical environments. PPDLS~\cite{PPDLS} focuses on tobacco leaves, while MSU-PID~\cite{MSU-PID}  tracks Arabidopsis plant growth. KOMATSUNA~\cite{KOMATSUNA}  captures images of Komatsuna leaves, all addressing the challenge of segmenting plants with diverse shapes, textures, and backgrounds.
\item \textbf{Scene} category includes datasets for scene understanding, featuring a range of environments and contexts. ADE20K~\cite{ADE20K} covers diverse scenes and objects, for instance segmentation, while LVIS~\cite{LVIS} focuses on long-tailed objects. COCO~\cite{COCO} is widely used for object detection and segmentation, with extensive image data. SUN~\cite{SUN} targets everyday scenes with semantic annotations, and Places~\cite{Places} focus on scene objects in various settings. Cityscapes~\cite{cityscapes} is a specialist in urban street scenes, offering panoptic segmentation data. NDISPark~\cite{NDISPark1,NDISPark2} and STREETS~\cite{streets} provide instance-level segmentation for parking lots and street-level videos, respectively. iShape~\cite{ishape} emphasizes indoor shape and structure, and IBD~\cite{IBD} focuses on individual cell structure segmentation. These datasets help address the challenges of segmenting complex scenes in diverse environments. 
\item \textbf{Egocentric} datasets are captured from a first-person perspective, often using wearable cameras. VISOR~\cite{VISOR} includes complex real-world scenes, while GTEA~\cite{gtea} and EgoHOS~\cite{EgoHos} focus on human-object interactions, tracking hand movements and object manipulation, with challenges such as occlusions and rapid motion.
\item \textbf{Paintings} datasets focus on the segmentation of artistic works. DRAM~\cite{DRAM} covers art paintings, while SegCLP~\cite{SegCLP} focuses on Chinese paintings. CLP~\cite{CLP} deals with Chinese landscape art, addressing the challenges in distinguishing artistic styles, brush strokes, and color variations in fine art.
\item \textbf{Others}: The ``Others'' category includes datasets that don not fall into the above categories but are still valuable for specialized tasks in image segmentation. This may include niche applications such as industrial inspection, satellite imagery, or other unique domains that require specific segmentation techniques. These datasets present a diverse range of challenges, depending on the application, and require highly specialized models to achieve accurate segmentation results. \textbf{Others} encompasses diverse datasets for specialized image segmentation tasks. BBBC038v1~\cite{BBBC038v1} targets biological nuclei segmentation, while DOORS~\cite{DOORS} focuses on doorways and architectural elements. TimberSeg~\cite{TimberSeg} is dedicated to operator-view timber log segmentation, and OVIS~\cite{OVIS} handles occlusion in videos. Hypersim~\cite{hypersim} offers photorealistic indoor masks for segmentation, while WoodScape~\cite{woodscape} focuses on surround fisheye views. Plittersdorf~\cite{Plittersdorf} provides data for wildlife trap segmentation, and ZeroWaste-f~\cite{zerowaste} focuses on recycling waste. These datasets cater to unique, niche applications requiring specialized segmentation techniques across various domains.
\end{itemize}

\subsubsection{Medical datasets}

Medical scene datasets encompass various imaging modalities (such as CT, PET, MRI, ultrasound, etc.), different body parts (such as head and neck, chest, abdomen, etc.), and diverse dimensions (2D and 3D). To better understand and analyze these datasets, we classify them into single-organ and multi-organ categories. In the single-organ category, we further refine the classification based on specific body parts.

\begin{table*}
\renewcommand{\arraystretch}{1.2}

\centering
\caption{Single-organ datasets, including the dataset name, year of publication, related papers, imaging modalities, the number of images and masks, data dimensions, and descriptive annotations.}
\label{tab:single-organ}
\begin{tabular}{>{\centering\arraybackslash}p{1.5cm}>{\centering\arraybackslash}p{2.2cm}>{\centering\arraybackslash}p{0.9cm}>{\centering\arraybackslash}p{1cm}cccc>{\centering\arraybackslash}p{3.5cm}}
\toprule
 Anatomical&Dataset & Year& Pub.& Modality& Images &Masks &Dimension& Descr.\\
 \midrule
  \multirow{5}{*}{Head and Neck}&BraTs~\cite{BraTs}& 2014& TMI& MRI& -& 12591& 3D& Multimodal Brain Tumor\\
 & ANDI~\cite{ANDI}& 2015& -& MRI& 68& -& 3D&Hippocampal segmentation\\
  & Hippo~\cite{MSD}& 2019& Arxiv& MRI& 260& -& 3D&Hippocampus\\
  & Hecktor~\cite{hecktor}& 2022& MIA& PET/CT& 489& 489& 3D& Head and neck\\
& TN3K~\cite{TCK}& 2023& CBM& Ultrasound& 3493& -& 2D&thyroid nodules\\
\specialrule{0em}{1pt}{1pt}
\hdashline
\specialrule{0em}{1pt}{1pt}
\multirow{9}{*}{Thorax}
 & Chest-Xray~\cite{chest-Xray}& 2014& QIMS& X-ray& -&  704& 2D& Chest X-ray\\
 & Covid19~\cite{covid19}& 2019& CBM& X-ray& 11956& -& 2D&Pulmonary tissues\\
 & Breast-US~\cite{USBreast}& 2020& SCI& Ultrasound& -&  630& 2D& Breast Ultrasound\\
 & BUID~\cite{buid}& 2020& dib& Ultrasound& 780& 529&2D& Breast tumor\\
 & LA~\cite{LA}& 2021& MIA& LGE-MRI& 154&- &3D& Left atrium\\
 & CIR~\cite{CIR}& 2022& MICCAI& CT& 956&-&3D& Lung nodule\\
 & ACDC~\cite{fewshot-sam2}& 2024& Arxiv& MRI, CMR& 1808& - & 2D& Cardiac Image\\
& CMR T1-Map~\cite{fewshot-sam2}& \raisebox{-0.5\height}{2024}& \raisebox{-0.5\height}{Arxiv}& \raisebox{-0.5\height}{CMR}& \raisebox{-0.5\height}{50}&\raisebox{-0.5\height}{-}& \raisebox{-0.5\height}{2D}& \raisebox{-0.5\height}{Myocardium Image}\\

\specialrule{0em}{1pt}{1pt}
\hdashline
\specialrule{0em}{1pt}{1pt}
 \multirow{4}{*}{Abdomen}
& Pancreas~\cite{Pancreas}& 2018& ASO& CT& 285& -& 3D& Pancreatic parenchyma\\
& Kvasir~~\cite{Kvasir}& 2020& MMM& Colonoscopy& 1000& -&2D& Polyps\\
& Kidney-US~\cite{ct2us}& 2022& Ultrasonics& Ultrasound& -&  4586& 2D&CT2US for Kidney\\
&LiTs~\cite{ctliver2023}& 2023& MIA& CT&  -& 5501&3D& Liver tumor\\

 \specialrule{0em}{1pt}{1pt}
\hdashline
\specialrule{0em}{1pt}{1pt}
  \multirow{3}{*}{Pelvis}& Prostate~\cite{Prostate}& 2020& TMI& MRI& 79& -& 3D& Prostate segmentation \\
 & MMOTU~\cite{Ovarian}& 2022& Arxiv& Ultrasound& -&  1469& 2D& Ovarian Tumor\\
 & OAI-ZIB~\cite{novel-sam2knee-zeroshot}& 2024& Arxiv& MRI& 488& -& 3D& Knee MR volumes\\
 \specialrule{0em}{1pt}{1pt}
\hdashline
\specialrule{0em}{1pt}{1pt}
 \multirow{3}{*}{Bone and spine}& Spine-MRI~\cite{MRI-spine}& 2017& Neuroimage& MRI& -& 551& 3D& Spinal Cord Grey Matter\\
 & SpineSeg~\cite{spineseg}& 2020& TMI& MR& 215& -&3D& SpineMR\\
  & VerSe~\cite{VerSe}& 2021& MIA& CT& 374&  4505&3D&Vertebrae Segmentation\\
 \specialrule{0em}{1pt}{1pt}
\hdashline
\specialrule{0em}{1pt}{1pt}
 \multirow{3}{*}{Lesion}
  & ISIC~\cite{ISIC}& 2017 & ISBI& Dermoscopy& 2750& -&  2D& Skin Lesion\\
 & fdg-PET~\cite{PET/CT-LESION}& 2022& Nature& PET/CT& -& 1015& 3D& Annotated tumor lesions\\
 & ISLES\cite{ISLES22,isles18}& 2022& SCI& MRI& 400& -& 3D&Stroke Lesion\\
 \specialrule{0em}{1pt}{1pt}
\hdashline
\specialrule{0em}{1pt}{1pt}
\multirow{7}{*}{Large Scale}
 & SA-1B~\cite{SAM}& 2023& ICCV& -& 11M&  1.1B & 2D&Universal segmentation
\\
 & SA-Med2D~\cite{Sa-med2d-20m}& 2023& Arxiv& 10 modalities& 4.6M&  19.7M &2D&IMIS, Zero-shot and so on
\\
 & MedSAM~\cite{MedSAM}& 2023& Nature& 10 modalities& 1570263&  1570263 & 2D, 3D&Universal segmentation
\\
 & COSMOS~\cite{cosmos}& 2024& MIA& 18 modalities& 1050K&  6033K & 2D&IMIS, Zero-shot and so on\\
 & onePrompt~\cite{onePrompt}& 2024& CVPR& -& -& ~3000 & 2D, 3D& IMIS \\ 
 & OmniMedVQA~\cite{omnimedvqa}& 2024& CVPR& 12 modalities& 118010& -& 2D, 3D& 20 anatomical regions\\
 & IMed-361M~\cite{IMIS}& 2024& BIBM& 14 modalities& 6.4M&  361M & 2D, 3D&IMIS, Zero-shot\\
 \specialrule{0em}{1pt}{1pt}
\hdashline
\specialrule{0em}{1pt}{1pt}
\multirow{6}{*}{Other}
& GlaS~\cite{glas}& 2017& MIA& H\&E stained images& 165& -& 2D&Colorectal Adenocarcinoma\\
& HC~\cite{HC}& 2018& PIoS& Ultrasound& 999& -& 2D&Fetal head circumference\\
& CRAG~\cite{CRAGData}& 2019& MIA& H\&E stained images& 213& -& 2D&Adenocarcinoma Slides\\
 & OCTA-500~\cite{OCTA500}& 2020& MIA& OCT/OCTA& 500& -& 3D &Retinal vessels\\
& EBHI~\cite{ebhi}& 2023& FIM.& H\&E stained images& 4456& -& 2D&Histopathological Images\\
& FUSC~\cite{fuseg}& 2024& MDPI& RGB& 1210& -& 2D&Foot ulcer\\
\bottomrule
\end{tabular}
\end{table*}

Table \ref{tab:single-organ} categorizes the single-organ datasets into the following anatomical regions: Head and Neck, Thorax, Abdomen, Pelvis, Bone and Spine, Lesion, and Others. In addition, we have also collected large-scale datasets.

\begin{itemize}
\item \textbf{Head and Neck}: BraTs~\cite{BraTs}, a multimodal MRI dataset with 12,591 images, focused on brain tumor research; Hecktor ~\cite{hecktor}, a PET/CT dataset with 489 images, primarily used for head and neck tumor analysis; Hippo~\cite{MSD} contains 260 MRI images focusing on hippocampal region segmentation; ANDI~\cite{ANDI}, a dataset with 68 MRI images for hippocampal segmentation; and TN3K~\cite{TCK}, an ultrasound dataset with 3,493 images, focusing on thyroid nodule analysis.
\item \textbf{Thorax}: Breast-US~\cite{USBreast}, a breast ultrasound dataset comprising 630 images; Chest-Xray~\cite{chest-Xray}, which includes 704 X-ray images for chest X-ray analysis; Covid19~\cite{covid19}, containing 11,956 X-ray images, primarily used for lung tissue analysis; ACDC~\cite{fewshot-sam2}, a dataset with 1,808 MRI and CMR images, focused on cardiac imaging; and CMR T1-Map~\cite{fewshot-sam2}, a myocardial imaging dataset with 50 CMR images. LA~\cite{LA}, a Left Atrium segmentation dataset utilizing LGE-MRI, with 154 3D images; CIR~\cite{CIR}, a lung nodule detection dataset from MICCAI, consisting of 956 3D CT images; and BUID~\cite{buid}, a breast tumor detection dataset based on ultrasound, containing 780 images, 529 of which are annotated in 2D. 
\item \textbf{Abdomen}: Kidney-US~\cite{ct2us}, which contains 4,586 ultrasound images for kidney analysis in CT2US tasks; LiTs~\cite{ctliver2023}, a dataset of 5,501 CT images aimed at liver tumor analysis; Kvasir~\cite{Kvasir}, featuring 1,000 endoscopic images designed for colon polyp detection; and \textbf{Pancreas}~\cite{Pancreas}, a dataset with 285 CT images dedicated to pancreatic tissue analysis.
\item \textbf{Pelvis}: Prostate~\cite{Prostate}, a prostate segmentation dataset with 79 3D MRI images; MMOTU~\cite{Ovarian}, which contains 1,469 ultrasound images for ovarian tumor detection; and OAI-ZIB~\cite{novel-sam2knee-zeroshot}, a knee MR dataset with 488 3D MRI images.
\item \textbf{Bone and Spine}: Spine-MRI~\cite{MRI-spine}, a dataset designed for spinal cord gray matter detection with 551 3D MRI images; VerSe~\cite{VerSe} , a vertebrae segmentation dataset comprising 374 3D CT images and 4,505 masks; and SpineSeg~\cite{spineseg}, a spine MR imaging dataset containing 215 3D images.
\item \textbf{Lesion}: Fdg-PET~\cite{PET/CT-LESION}, a tumor lesion dataset with 1,015 3D PET/CT images; ISIC~\cite{ISIC}, which contains 2,750 2D images for skin lesion detection; and ISLES~\cite{isles18,ISLES22}, a dataset for stroke lesion detection with 400 3D MRI images.
\item \textbf{Large Scale}: IMed-361M~\cite{IMIS}, a large-scale dataset with 6.4 million images across 14 modalities, supporting zero-shot segmentation; SA-1B~\cite{SAM}, a universal segmentation dataset with 11 million images spanning 2D modalities; COSMOS~\cite{cosmos}, an 18-modality dataset with 1.05 million images, designed for zero-shot segmentation; onePrompt~\cite{onePrompt}, which contains approximately 3,000 2D and 3D images supporting IMIS; SA-Med2D~\cite{Sa-med2d-20m}, a 2D universal segmentation dataset with 4.6 million images across 10 modalities; MedSAM~\cite{MedSAM}, a universal segmentation dataset with 1.57 million images across 10 modalities; and OmniMedVQA~\cite{omnimedvqa}, a medical question-answering dataset encompassing 12 modalities and 118,010 images.
\end{itemize}
\begin{itemize}
    \item \textbf{Others}:  HC~\cite{HC}, a dataset of 999 2D ultrasound images for fetal head circumference measurement; OCTA-500~\cite{OCTA500}, which includes 500 3D images for retinal vessel detection; EBHI~\cite{ebhi}, a dataset with 4,456 2D H\&E stained histopathological images; CRAG~\cite{CRAGData}, which contains 213 2D H\&E stained adenocarcinoma slides; GlaS~\cite{glas}, a dataset with 165 2D H\&E stained images for colorectal adenocarcinoma detection; and FUSC~\cite{fuseg}, a foot ulcer detection dataset with 1,210 2D RGB images.
\end{itemize}

\begin{table*}
\renewcommand{\arraystretch}{1.2}

\centering
\caption{Collection of multi-organ datasets, detailing publication year, source, imaging modality, dimensionality, and the specific anatomical regions covered.}
\label{tab:multi-organ}
\begin{tabular}{cccccc}
\toprule
 Dataset & Year& Pub.& Modality& Dimension&Anatomical\\
 \midrule
  \multirow{2}{*}{Dense-Vnet~\cite{dense-vnet}}& \multirow{2}{*}{2018}& \multirow{2}{*}{TMI}& \multirow{2}{*}{CT, MRI}& \multirow{2}{*}{3D}& pancreas, gastrointestinal tract (esophagus, \\
 & & & & &  stomach, duodenum), liver, spleen, left kidney, gallbladder\\
   \specialrule{0em}{1pt}{1pt}
\hdashline
\specialrule{0em}{1pt}{1pt}
  \multirow{2}{*}{MSD~\cite{MSD}}& \multirow{2}{*}{2019}& \multirow{2}{*}{Arxiv}& \multirow{2}{*}{7 modalities}& \multirow{2}{*}{2D, 3D}& BrainTumour, Heart, Liver, Hippocampus,\\
 & & & & &Prostate, Lung, Pancreas, HepaticVessel, Colon, Spleen\\
  \specialrule{0em}{1pt}{1pt}
\hdashline
\specialrule{0em}{1pt}{1pt}CT-organ~\cite{ctOrgan}& 2019& TCIA& CT& 3D&Liver, bladder, lungs, kidney, and bone\\
  \specialrule{0em}{1pt}{1pt}
\hdashline
\specialrule{0em}{1pt}{1pt} BTCV~\cite{BTCV}& 2020& TMI& CT& 3D&Abdomen, Cervix, Brain, Heart, Canine Leg\\
   \specialrule{0em}{1pt}{1pt}
\hdashline
\specialrule{0em}{1pt}{1pt} \multirow{2}{*}{AbdomenCT-1K~\cite{Abdomenct-1k}}& \multirow{2}{*}{2020}& \multirow{2}{*}{TPAMI}& \multirow{2}{*}{CT}& \multirow{2}{*}{3D}& Multi-center Abdominal Organ: \\
 & & & & &Liver, kidney, spleen, and pancreas.\\
  \specialrule{0em}{1pt}{1pt}
\hdashline
\specialrule{0em}{1pt}{1pt} CHAOS~\cite{chaos}& 2021& MIA& CT, MRI& 2D, 3D& Liver, kidney(s), spleen\\
  \specialrule{0em}{1pt}{1pt}
\hdashline
\specialrule{0em}{1pt}{1pt} \multirow{2}{*}{Flare~\cite{Flare}}& \multirow{2}{*}{2022}& \multirow{2}{*}{Arxiv}& \multirow{2}{*}{CT}& \multirow{2}{*}{3D}& Abdominal diseases: liver kidney, spleen,
\\
 & & & & &pancreatic, stomach  sarcomas, colon ovarian, and bladder.\\
  \specialrule{0em}{1pt}{1pt}
\hdashline
\specialrule{0em}{1pt}{1pt} \multirow{3}{*}{AMOS~\cite{AMOS}}& \multirow{3}{*}{2022}& \multirow{3}{*}{NIPS}& \multirow{3}{*}{CT, MRI}& \multirow{3}{*}{3D}& Spleen, right kidney, left kidney, gallbladder, esophagus, liver,\\
 & & & & & stomach, aorta, inferior vena cava, pancreas, right adrenal gland,\\
 & & & & &left adrenal gland, duodenum, bladder, prostate/uterus.\\
 \specialrule{0em}{1pt}{1pt}
\hdashline
\specialrule{0em}{1pt}{1pt} TotalSegmentatorV2-CT~\cite{totalsegmentatorCT}& 2023& RAI& CT, MRI& 3D& 27 organs, 59 bones, 10 muscles, 8 vessels.\\
\bottomrule
\end{tabular}
\end{table*}

Table \ref{tab:multi-organ} compiles multi-organ datasets that encompass multiple body organs, modalities, and dimensions.

\begin{itemize}
\item The MSD dataset~\cite{MSD} provides a comprehensive collection for multi-organ segmentation with 7 modalities, including both 2D and 3D images. It covers a wide range of organs such as Brain Tumour, Heart, Liver, Hippocampus, and Spleen, offering a valuable benchmark for evaluating and developing segmentation algorithms across multiple organ types and imaging modalities.
\item CT-organ~\cite{ctOrgan} is a 3D CT dataset focusing on abdominal organs, including liver, bladder, lungs, kidneys, and bones. This dataset plays a crucial role in advancing CT-based segmentation techniques, particularly for abdominal imaging, and serves as a resource for improving segmentation performance in clinical and research settings.
\item BTCV~\cite{BTCV} is a dataset designed for the segmentation of body tumors and various organs. It includes challenging cases of diverse organ structures and tumors, providing a platform for evaluating segmentation methods on real-world medical images, especially in the context of tumor detection and organ delineation in the abdomen and thoracic regions.
\item AbdomenCT-1K~\cite{Abdomenct-1k} includes 3D CT images from multiple medical centers, offering a large-scale resource for abdominal organ segmentation. With a wide variety of images from different scanners, it provides an opportunity to test and refine segmentation methods that generalize across clinical settings, making it a critical tool for training deep learning models.
\item CHAOS~\cite{chaos} consists of both 2D and 3D CT/MRI images focusing on abdominal organs, including liver, kidneys, and spleen. The dataset was specifically designed for segmentation algorithm development, providing ground truth annotations and diverse examples that facilitate the training and evaluation of models for abdominal organ segmentation in both CT and MRI modalities.
\item Flare~\cite{Flare} offers 3D CT images of 13 abdominal organs, such as liver, kidneys, spleen, and intestines. The dataset is intended to support research in the segmentation of abdominal organs, with its focus on CT images helping to advance the development of segmentation techniques that can handle multiple organ types in a single scan.
\item AMOS~\cite{AMOS} is a 3D CT and MRI dataset containing images of 15 abdominal organs. This dataset supports multi-organ segmentation research, enabling the development of algorithms that can segment multiple organs simultaneously. It is particularly valuable for testing deep learning models in the complex task of segmenting multiple structures within the abdomen using both CT and MRI scans.
\item TotalSegmentatorV2-CT~\cite{totalsegmentatorCT} provides 3D CT and MRI images for whole-body organ segmentation. This dataset focuses on segmenting all major organs in the human body and is designed to aid in the development of full-body segmentation algorithms, providing a valuable resource for researchers working on multi-organ segmentation in both CT and MRI modalities.
\item Dense-Vnet~\cite{dense-vnet} includes 3D CT and MRI images and focuses on abdominal organs such as the pancreas, liver, spleen, and gastrointestinal structures. The dataset supports research into dense organ segmentation techniques, providing a rich source of images for testing models that aim to segment complex anatomical structures in the abdominal region.
\end{itemize}

\subsection{Evaluate Metrics}

To effectively measure the performance and effectiveness of segmentation algorithms, evaluation metrics are crucial. These metrics not only help us verify whether the segmentation results meet expectations, but also provide objective standards for comparing the performance of different algorithms and models. In image segmentation tasks, the evaluation primarily focuses on two aspects: first, the \textbf{accuracy of the segmentation results}, i.e., whether the segmentation can accurately capture the target region, especially when the target's shape is complex or there is overlap; and second, the \textbf{robustness of the model}, i.e., how well the model can adapt to different input conditions, such as changes in lighting, occlusions, and other factors. By considering these two factors comprehensively, we can fully assess the model's practical performance and provide guidance for subsequent optimization.

\subsubsection{Intersection over Union (IoU)}

\textbf{IoU} is a metric used to measure the overlap between the predicted segmentation region and the ground truth region. It evaluates the accuracy of the segmentation by calculating the ratio of the intersection to the union of the predicted and true regions. A higher IoU indicates better segmentation accuracy.
\begin{equation}
\text{IoU} = \frac{|A \cap B|}{|A \cup B|},
\end{equation}
\( A \) is the predicted segmentation region, \( B \) is the ground truth region, \( |A \cap B| \) is the area of intersection between the predicted and ground truth regions, \( |A \cup B| \) is the area of union between the predicted and ground truth regions.

\subsubsection{Dice Similarity Coefficient (Dice)}

The Dice coefficient is another widely used metric similar to IoU, but mathematically different. It is particularly useful for assessing similarity between two sets, especially when the target region is small or irregular, as is common in medical image segmentation.
\begin{equation}
\text{Dice} = \frac{2 \cdot |A \cap B|}{|A| + |B|},
\label{eq:2}
\end{equation}
The elements in \eqref{eq:2} are the same as in the IoU formula.

\subsubsection{Mean Intersection over Union (mIoU)}

mIoU is the average IoU across all classes in a multi-class segmentation task. In multi-class problems, we compute the IoU for each class and then average them. mIoU provides a comprehensive measure of performance across all classes in a segmentation task.
\begin{equation}
    \text{mIoU} = \frac{1}{N} \sum_{i=1}^{N} \frac{|A_i \cap B_i|}{|A_i \cup B_i|},
\end{equation}
\( N \) is the total number of classes, \( A_i \) and \( B_i \) are the predicted and ground truth regions for class \( i \), \( |A_i \cap B_i| \) and \( |A_i \cup B_i| \) are the intersection and union areas for class \( i \).

\subsubsection{Pixel Accuracy (PA)}

Pixel Accuracy is the simplest segmentation metric, which measures the proportion of correctly predicted pixels over the total number of pixels. PA can be used to evaluate the overall performance of the model, but may not reflect true performance in cases of class imbalance.
\begin{equation}
    \text{PA} = \frac{1}{N} \sum_{i=1}^{N} 1 (y_i = \hat{y}_i).
\end{equation}
\( y_i \) is the ground truth value of the \( i \)-th pixel, \( \hat{y}_i \) is the predicted value of the \( i \)-th pixel, 1 is the indicator function, which is 1 if \( y_i = \hat{y}_i \) (the prediction is correct), and 0 otherwise, \( N \) is the total number of pixels.
 \section{SAM2 for Video} \label{Video}

In this section, we will summarize the video segmentation architectures based on SAM and SAM2, and compare them with other state-of-the-art architectures. In addition, we will introduce relevant video segmentation datasets and provide commonly used evaluation metrics, along with an analysis of segmentation performance.

\subsection{Video Segmentation Networks}

From cutting-edge model architectures to the continuous expansion of application scenarios, video segmentation technology has demonstrated strong potential for handling complex temporal data and long-duration video segmentation tasks. Traditional video segmentation tasks are typically classified into video object segmentation (VOS), video semantic segmentation (VSS), and video instance segmentation (VIS). 

With the emergence of general-purpose models, the concepts of zero-shot and few-shot segmentation have introduced new directions for video segmentation. In response to these complex video segmentation tasks, models based on various architectures, such as STCN, SegGPT, and DeAOT, have successfully addressed challenges including sVOS and iVOS by using advanced mechanisms and algorithms. Meanwhile, SAM2 and its derivative versions have made significant progress in integrating interactive segmentation, memory tracking, and few-shot learning techniques. In the following, we will provide a detailed introduction from three perspectives in Table \ref{video-networks}: \textbf{state-of-the-art techniques}, \textbf{SAM-based} models, and \textbf{ SAM2-based} models, followed by an in-depth analysis.

\begin{table*}
\renewcommand{\arraystretch}{1.2}

\centering
\caption{Video segmentation networks, categorizing them by their methodological approaches, publication years, backbone architectures, distinctive features, and the tasks they address, including state-of-the-art, SAM-based and SAM2-based methods.}
\label{video-networks}
\begin{tabular}{>{\raggedright\arraybackslash}p{2cm}@{}cccc>{\centering\arraybackslash}p{4cm}@{}c}
\toprule
 Classification&Method& Year& Pub. & Backbone& Feature&  Tasks\\
 \midrule
  \multirow{9}{*}{State of the Arts}& STCN~\cite{STCN}& 2021 & NIPS & ResNet -50& Diversified Voting&sVOS\\
 & SegGPT~\cite{Seggpt}& 2021& ICCV& Vision Transformer& Mask Prompting&sVOS, iVOS, Zero-shot\\ 
 & DeAOT~\cite{DeAOT}& 2022& NIPS & Vision Transformer& Decoupled Propagation&sVOS\\ 
  & RDE~\cite{RDE}& 2022 & CVPR & STM~\cite{STM}& Self-correction&sVOS\\ 
  & XMem~\cite{Xmem}& 2022 & ECCV & ResNet-50& Memory consolidation&sVOS\\ 
  & DEVA~\cite{DEVA}& 2023& ICCV& Mask2FormerR50& Decoupled segmentation&Few-shot\\ 
  & DDMemory~\cite{DDmemory2023Lvos}& 2023 & ICCV & ResNet& Dynamicmemory banks&LVOS\\ 
& Cutie~\cite{Cutie}& 2024 & CVPR & ResNet- 50/Res-net 18& Object-level memory &sVOS, LVOS\\ 
& LiVOS~\cite{LiVos}& 2024 & Arxiv & ResNet-50 & Lightweight memory&sVOS\\ 
\specialrule{0em}{1pt}{1pt}
\hdashline
\specialrule{0em}{1pt}{1pt}
\multirow{10}{*}{SAM-based}& \raisebox{-0.5\height}{SAM~\cite{SAM}}& \raisebox{-0.5\height}{2023}& \raisebox{-0.5\height}{ICCV}& \raisebox{-0.5\height}{Vision Transformer}& Universal Interactive Segmentation& \raisebox{-0.5\height}{sVOS, iVOS, Zero-shot}\\
&DEVA-SAM\cite{DEVA}& 2023& ICCV& SAM backbone& Decoupled Propagation& few-shot\\
&SAM-PT~\cite{sam-pt}& 2023& Arxiv& SAM backbone& Point Propagation&iVOS, Zero-shot\\
&SurgicalSAM~\cite{surgicalsam}& 2023& AAAI& SAM backbone& Prototype-tuning&Zero-shot\\
&MemSAM~\cite{memsam}& 2024& CVPR& MedSAM~\cite{MedSAM} backbone& Spatiotemporal Memory& sVOS\\
&VideoSAM~\cite{videosam}& 2024& Arxiv& SAM backbone& Memory Tracking& sVOS\\
&SAM-PD~\cite{sampd}& 2024& Arxiv& SAM backbone& Prompt Denoising&sVOS, iVOS\\
&RAP-SAM~\cite{rapSam}& 2024& Arxiv& SAM backbone& Real-time All-purpose&VSS, VIS, VPS, iVOS\\

\specialrule{0em}{1pt}{1pt}
\hdashline
\specialrule{0em}{1pt}{1pt}
\multirow{8}{*}{SAM2-based}&SAM2~\cite{SAM2} & 2024& Arxiv& Vision Transformer& Universal&sVOS, iVOS, Zero-shot\\
&Yolo-SAM2~\cite{yoloSam2}& 2024& Arxiv& YOLO v8 + SAM2& Bounding box-based &Zero-shot\\
  &SurgicalSAM2~\cite{surgicalSam2}& 2024& Arxiv& SAM2 backbone& Efficient Frame Pruning&Zero-shot(Real time)\\
&PolypSAM2~\cite{polypSAM2}& 2024& Arxiv& SAM2 backbone& different prompts&Zero-shot\\
&SAMWISE~\cite{samwise}& 2024& Arxiv& SAM2 backbone& Multimodal Temporal&RVOS\\
&Det-SAM2~\cite{detsam2}& 2024& Arxiv& SAM2 backbone& Pipeline &LVOS\\
&SAMURAI~\cite{yang2024samurai}& 2024& Arxiv& SAM2 backbone& Robust & Zero-shot\\

  &SAM2Long~\cite{sam2long}& 2024& Arxiv& SAM2 backbone& Pathway-optimization&sVOS, iVOS, Zero-shot\\
 \bottomrule
\end{tabular}
\end{table*}

\subsubsection{State of the Arts}

The state-of-the-art models in video segmentation have significantly improved the ability to handle dynamic changes and long-duration video data by leveraging innovative deep learning technologies and architectures. Next, we will dive into the core features of these cutting-edge models and their notable contributions to video segmentation tasks.

\begin{itemize}
\item STCN~\cite{STCN} utilizes a ResNet-50 architecture and diverse voting across multiple frames to enhance video object segmentation. By integrating spatio-temporal information, STCN captures the dynamic changes and spatial relationships of objects, making it well suited for complex tasks like sVOS and iVOS, where objects undergo temporal and spatial variations.
\item SegGPT~\cite{Seggpt} employs Vision Transformer (ViT) and Mask Prompting to handle complex objects in video segmentation. The ViT captures long-range dependencies across frames, while Mask Prompting improves segmentation accuracy, making SegGPT particularly effective in tasks like video object segmentation and instance segmentation, where precise segmentation is crucial.
\item DeAOT~\cite{DeAOT} introduces decoupled propagation mechanisms, which enhance flexibility in handling dynamic target movement and scene changes. This mechanism improves object tracking accuracy, making DeAOT especially suitable for long-duration video segmentation tasks like iVOS and sVOS, where complex object motion must be tracked over time.
\item RDE~\cite{RDE} enhances segmentation accuracy with region-based dynamic encoding, combining local and global contextual information to solve issues like occlusions and object interactions. RDE is robust in long-duration video segmentation tasks and performs efficiently, especially in real-time applications, where quick response and accuracy are critical.
\item XMem~\cite{Xmem} leverages memory consolidation techniques to improve long-term object tracking. By storing and updating target representations across frames, XMem ensures consistent tracking even with occlusions or appearance changes. This makes XMem ideal for long video sequences, particularly in tasks such as sVOS and iVOS that require identity maintenance over time.
\item DEVA~\cite{DEVA} uses the Mask2FormerR50 architecture for few-shot decoupled segmentation, enhancing generalization to new object categories with limited labeled data. This few-shot learning capability makes DEVA highly effective for tasks with sparse data, excelling in video object segmentation where only a small number of samples are available for training.
\item DDMemory~\cite{DDmemory2023Lvos} employs dynamic memory banks to handle long-duration video object segmentation, storing, and updating target representations across multiple frames. This method improves segmentation precision, making it ideal for addressing challenges such as occlusion and motion blur in long-term video segmentation tasks.
\item Cutie~\cite{Cutie} optimizes video object segmentation using a lightweight, object-level memory architecture. This design balances computational efficiency and segmentation performance, making it well suited for real-time applications, particularly in resource-constrained environments where performance and efficiency are crucial for video segmentation tasks.
\item LiVOS~\cite{LiVos} adopts a lightweight memory architecture to optimize video object segmentation, maintaining both segmentation accuracy and computational efficiency. This model is particularly effective for long-duration video segmentation, ensuring high-precision target identification and continuous tracking, even as the target changes over time.

\end{itemize}

\subsubsection{SAM-based Models}

This class of models based on the SAM architecture represents the latest advancements in video segmentation, aiming to enhance performance in interactive segmentation, memory tracking, and few-shot learning tasks. SAM and its derivative models exhibit excellent performance in handling dynamic video scenes and long temporal sequences by incorporating innovations such as spatio-temporal memory and decoupled propagation mechanisms. Next, we will provide a detailed overview of how these models improve the accuracy and efficiency of video object segmentation through various techniques.

\begin{itemize}
\item SAM~\cite{SAM} is a versatile interactive segmentation model based on Vision Transformer (ViT), capable of handling sVOS, iVOS, and zero-shot segmentation without labeled data. It captures long-range dependencies, making it ideal for dynamic segmentation tasks.
\item MemSAM~\cite{memsam} extends SAM with spatio-temporal memory, improving object tracking across frames. This enhancement aids in long-duration video segmentation, particularly for sVOS and iVOS tasks, by retaining and updating object representations.
\item DEVA-SAM~\cite{DEVA} introduces decoupled propagation mechanisms, which allow SAM to handle complex scenes with significant object changes. This improves flexibility and robustness for long-duration video segmentation tasks.
\item VideoSAM~\cite{videosam} enhances video segmentation by focusing on dynamic memory tracking. It ensures accurate segmentation across frames, even in challenging scenarios like occlusions, by maintaining object identities over time.
\item SAM-PT~\cite{sam-pt} improves SAM’s segmentation by using point propagation, leveraging sparse point annotations to refine object boundaries. This technique enhances accuracy in tasks like object instance and video object segmentation, where precision is crucial.
\item SAM-PD~\cite{sampd} enhances segmentation by reducing noise in prompts. This approach improves accuracy in ambiguous or noisy tasks, such as zero-shot segmentation and complex video sequences, ensuring focus on relevant features.
\item RAP-SAM~\cite{rapSam} is a real-time, versatile video segmentation model. It efficiently handles various challenges, from object tracking to scene understanding, suitable for interactive and live video analysis.
\item SurgicalSAM~\cite{surgicalsam} improves segmentation by combining multimodal temporal modeling with prototype tuning. This allows adaptation to unseen categories, making it effective for medical imaging tasks, particularly in surgery, where precise segmentation is critical.
\end{itemize}

\subsubsection{SAM2-based Models}

The SAM2-based models represent a significant advancement in the field of real-time video segmentation, leveraging the improved SAM2 architecture to address a variety of challenges. These models combine powerful techniques, such as efficient frame pruning, multimodal temporal modeling, and pathway optimization, to enhance segmentation accuracy and speed. 
Designed for applications that require real-time processing and adaptability, SAM2-based models excel in tasks like zero-shot segmentation, dynamic object tracking, and long-duration video analysis. In the following, we explore the key contributions of these models, highlighting their capabilities and applications in different domains.

\begin{itemize}
\item Yolo-SAM2~\cite{yoloSam2} integrates YOLO v8 and SAM2, focusing on enhancing object segmentation through bounding boxes. It achieves faster, accurate zero-shot segmentation, especially in real-time scenarios. Leveraging YOLO's detection and SAM2's segmentation, it addresses challenges in dynamic environments, offering a robust solution for tasks with limited or no labeled data.
\item SurgicalSAM2~\cite{surgicalSam2} improves the real-time segmentation performance of SurgicalSAM by utilizing efficient frame pruning techniques. This enables faster processing of video frames, making it suitable for live-streaming scenarios. Its precision and speed make it ideal for high-stakes applications like surgery and medical imaging, ensuring accurate results with minimal computational resources.
\item PolypSAM2~\cite{polypSAM2} examines SAM2's performance in polyp segmentation under various prompt settings, analyzing its strengths and limitations across datasets. The study evaluates segmentation accuracy, computational efficiency, and adaptability under different conditions, providing insights into the model's effectiveness in medical imaging tasks requiring high precision and reliability.
\item SAMWISE~\cite{samwise} introduces a novel real-time video segmentation method combining multimodal temporal modeling and prototype tuning. It adapts to new object categories and complex scenes without retraining, excelling in zero-shot segmentation tasks. Its efficiency and accuracy make it ideal for medical diagnostics and autonomous vehicles, where rapid adaptation to new data is critical.
\item Det-SAM2~\cite{detsam2}'s strengths lie in its automated object prompt generation and resource efficiency management, making it particularly suitable for long-duration video segmentation tasks requiring efficient inference.
\item SAMURAI~\cite{yang2024samurai} is an enhanced version of SAM2 designed for visual object tracking. By integrating temporal motion cues and motion-aware memory selection, it achieves robust, real-time tracking with accurate mask selection, excelling in zero-shot tasks without fine-tuning.
\item SAM2~\cite{SAM2} builds upon the original SAM model, offering enhanced efficiency and accuracy for real-time segmentation tasks. Powered by the Vision Transformer, SAM2 excels in dynamic video scenarios and extended tasks, delivering robust zero-shot capabilities for sVOS, iVOS, and applications with limited data. Its zero-shot inference demonstrates outstanding performance and resilience in surgical video segmentation~\cite{shen2024performance}.
\item SAM2Long~\cite{sam2long} optimizes SAM2 for long video segmentation using pathway strategies, maintaining accuracy over extended sequences. It handles longer temporal dependencies, ideal for video surveillance, environmental monitoring, and autonomous driving, requiring continuous object tracking.
\end{itemize}

\subsection{Video Datasets}

Table \ref{video-datasets} summarizes several important video segmentation datasets that have been developed for various segmentation tasks, including video object segmentation, motion tracking, and complex scene analysis. These datasets provide valuable resources for training and evaluating video segmentation models that address various real-world challenges. Each dataset varies in terms of the number of videos, object categories, annotations, and tasks, making them highly valuable for different research applications such as multi-object tracking, object detection, and real-time video segmentation. In the following, we briefly describe the key datasets and their respective characteristics and target tasks.

\begin{table*}[ht]
\renewcommand{\arraystretch}{1.2}

\centering
\caption{Detailed summary of video datasets, including the year of publication, publication venue, number of videos, frames, duration, objects, annotation type, and tasks.}
\label{video-datasets}
\begin{tabular}{@{}ccccccccl}
\toprule
Dataset& Year& Pub. & Videos& Frames&  Duration& Objects& Annotation &Tasks\\ \midrule
SegTrack~\cite{SegTrack}& 2012 & IJCV & 6 &      -&      -& 6 & 244  &Segmentation and Tracking\\
 SegTrack-v2~\cite{SegTrack-v2} & 2013 & ICCV & 14 & 976 & -& 24 & 1154  &Segmentation and Tracking\\
YouTube-Objects~\cite{YT-Objects} & 2013& CVPR & 126 & -& -& 96 & 2153  &Weakly-supervised learning\\
 FBMS-59~\cite{FBMS59} & 2014 & TPAMI & 59 & 13860 & 7.7(min)& 139 & 1465  &Motion Segmentation\\
Jump-cut~\cite{jumpcut}& 2015 & CVPR & 22 & -& -& 22 & 6331  &Video Cutout\\
DAVIS16~\cite{DAVIS16} & 2016 & CVPR & 50 & 3455 & 2.5(min)& 50 & 3455  &Video Object Segmentation\\
DAVIS17~\cite{DAVIS2017} & 2017 & arxiv & 90 & 10700 & 5.17(min) & 205 & 13543  &Video Object Segmentation\\
EndoVis17~\cite{EndoVis17} & 2017 & MICCAI & 8& 2040& -& 7& -& Robotic Instrument Segmentation\\
EndoVis18~\cite{EndoVis18} & 2018 & MICCAI & 10& 3000& -& 10& -& Instrument and Anatomical\\
YouTube-VOS~\cite{YT2019} & 2018 & ECCV & 4453 & 120532 & 334.8(min) & 7755 & 197272  & Video object segmentation\\
 CAMUS~\cite{CAMUS} & 2019 & TMI & 1000& -& -& -& -& Ultrasound\\
 VOT-LT 2019~\cite{VOT2019} & 2019 & ICCV & 50 & 215298 & 119(min) & 50 & 215298  & Object Tracking\\
  UVO~\cite{UVO} & 2021 & CVPR & 1200 & \( \sim \) 108000& 511(min) & 14748 & \( \sim \) 1327000  & open-world, class-agnostic\\
 VOT-LT 2022~\cite{VOT2022}& 2022 & ECCV & 52 & 168282 & 93(min) & 50 & 168282  & Object Tracking\\
 Endo NeRF~\cite{EndoNeRF} & 2022 & MICCAI& 6& 807& 0.6(min)& -& -& Surgical scene reconstruction\\
SurgToolLoc\cite{surgicalToolLoc} & 2022 & MICCAI & 24695& 1481700& 205.8(h)& 11& -& Surgical Tool Localization\\
VIPSeg~\cite{VIPSeg}	&2022 &CVPR &3536& -&353.6(min) & 124& 84750& Video Panoptic Segmentation\\
SUN-SEG~\cite{sun-seg}& 2022& MIR& 113& 158,690& 11.3(min)& -& -&Video polyp segmentation\\
 VOT-ST 2022~\cite{VOT2022} & 2022 & ECCV & 62 & 19903 & 11.1(min) & 62 & 19826  & Object Tracking\\
 VOT-ST 2023~\cite{VOT2023} & 2023 & ICCV & 144 & 298640 & 166(min) & 341 & -  & Tracking and Segmentation\\
 VOST~\cite{VOST} & 2023 & CVPR & 713 & 75547 & 252(min) & - & 175913  & Object transformation segmentation\\
 BURST~\cite{Burst} & 2023 & WACV & 2914 & 624240 & 1734(min) & 16089 & 600157  & Tracking and segmentation\\
MOSE\cite{Mose} & 2023 & ICCV & 2149 & \( \sim \)159600& 443.6(min) & 5200 & 431725  & Complex scene object segmentation\\
PUMaVOS~\cite{PuMaVOS+Xmem++} & 2023 & ICCV& 24& -&11.6(min) & -&21K  & Challenging segmentation\\
ESD~\cite{ESD}& 2023& Arxiv& 120&14563 & - & -& 14166& Event-based object segmentation \\	 	 
MeViS~\cite{MeVIS} & 2023& ICCV& 2006& -& -& 8171& 443K& Motion-guided Segmentation\\
PolypGen~\cite{polyp}& 2023& Sci.Data& 1537& 6500& -& -& 3762& Polyp detection\\
LVOS-V1~\cite{hong2023lvos} & 2023 & ICCV & 220 & 126280 & 351(min) & 282 & 156432& Long video segmentation\\
LVOS-V2~\cite{hong2024lvos} & 2024 & arxiv & 720 & 296401 & 823(min) & 1132 & 407945& Long video segmentation\\
SA-V~\cite{SAM2} & 2024 & CVPR & 50.9K& 4.2M& 196h & -& 642.6K& Multiple scenes and details\\
\bottomrule
\end{tabular}
\label{videoDataset} 
\end{table*}

\begin{itemize}
\item SegTrack\cite{SegTrack} is a small video segmentation dataset containing only 6 short videos with limited object categories, making it suitable for single-object segmentation tasks. SegTrack-v2\cite{SegTrack-v2} expands on this, featuring 14 videos to support multi-object segmentation tasks and improving annotation accuracy, serving as a key resource for early multi-object tracking research.
\item YouTube-Objects\cite{YT-Objects} includes 126 videos focusing on single-object segmentation, making it ideal for learning simple object segmentation from real-world scenarios. FBMS-59\cite{FBMS59} consists of 59 videos and 13,860 frames, covering 16 target categories, and is designed for evaluating multi-object segmentation. Jump-cut\cite{jumpcut} contains 22 videos emphasizing clip-based segmentation tasks, making it a high-quality resource for analyzing dynamic objects.
\item DAVIS16\cite{DAVIS16} and DAVIS17\cite{DAVIS2017} are key video segmentation datasets. DAVIS16 focuses on single-object segmentation, while DAVIS17 extends to multi-object segmentation with 90 videos and 13,543 frames. The dataset is known for high-quality annotations and complex scenes, providing a solid benchmark for video object segmentation tasks. 
\item EndoVis17\cite{EndoVis17}: A small dataset of 8 videos (255 frames each) for surgical instrument segmentation, with ground truth segmentation masks for each frame. EndoVis18\cite{EndoVis18}: Includes 19 sequences for surgical image segmentation, with 15 for training and 4 for testing. Each sequence contains 300 frames with complex annotations of anatomical structures and surgical tools.
\item CAMUS dataset\cite{CAMUS} includes data from 500 patients, with 450 used for training. It provides 900 echocardiography sequences from GE Vivid E95 scanners, featuring various image qualities and resolutions (584×354 to 1945×1181), lasting 10-42 seconds. 
\item Endo NeRF\cite{EndoNeRF}: Contains two surgical video clips, one with 63 frames and the other with 156 frames, focusing on endoscopic image analysis. SUN-SEG~\cite{sun-seg} is the first high-quality video polyp segmentation dataset, containing 158,690 colonoscopy video frames with various annotation types. SurgToolLoc\cite{surgicalToolLoc}: Comprising 24,695 30-second video clips at 60 fps, this dataset is used for endoscopic image analysis, with tool presence annotations and bounding box labels in the test set.
\item YouTube-VOS\cite{YT2019} is a large-scale dataset with 4,453 clips in 94 object categories, including humans, animals, and vehicles. It addresses spatio-temporal challenges and includes complex multi-object scenes, with 3,471 training and 507 validation videos for evaluating model performance.
\item VOT-LT Series \cite{VOT2019,VOT2022} is specialized for long-term video tracking tasks, containing 50 and 52 videos, respectively. It provides a large data volume with persistent targets, making it suitable for evaluating long-term stability and robustness. VOT-ST Series \cite{VOT2023,VOT2022} focuses on short-term tracking, with the 2023 version expanding to 144 videos and nearly 300,000 frames with finer annotations, supporting multi-object segmentation in dynamic scenes.
\item ESD~\cite{ESD} is the first 3D spatio-temporal event segmentation dataset, featuring 145 sequences and 21.88 million events for occluded object segmentation. VIPSeg~\cite{VIPSeg} is the first large-scale outdoor video panoptic segmentation dataset, containing 3,536 videos and 84,750 frames of pixel-level annotations.
\item VOST\cite{VOST} is a semi-supervised video object segmentation benchmark focused on complex object deformations. It includes targets that break, tear, or undergo significant shape changes. The dataset consists of over 700 high-resolution video clips, averaging 20 seconds each, with dense instance mask annotations. It is designed to capture the complete deformation process, offering challenging scenarios for VOS models.
\item UVO\cite{UVO} and BURST\cite{Burst} are large-scale multi-object segmentation datasets with 1,200 and 2,914 videos, respectively. UVO offers 14,748 objects with over 1.32 million annotations, focusing on undefined object segmentation. BURST covers 482 object categories and 600,157 annotations, serving as a critical resource for ultra-long video tasks.
\item MOSE\cite{Mose} is a video object segmentation (VOS) dataset for complex real-world scenarios, containing 2,149 video clips and providing 43,725 high-quality segmentation masks across multi-object scenes. The dataset is divided into 1,507 training videos, 311 validation videos, and 331 test videos, covering 36 categories with a total of 5,200 objects. PUMaVOS~\cite{PuMaVOS+Xmem++} is a video object segmentation dataset that covers partial and unusual objects, consisting of 24 video clips and 21K densely annotated frames. It focuses on partial objects (such as half faces, necks, tattoos) commonly retouched in film production, suitable for segmentation tasks in complex scenes.
\item MeViS~\cite{MeVIS} is a large-scale video segmentation dataset focused on motion expression-guided object segmentation. It contains 2,006 videos covering complex multi-object scenes, selected through rigorous visual and linguistic criteria, with detailed language expression annotations to advance motion-guided video segmentation research. PolypGen~\cite{polyp} is a large-scale dataset with 3,762 annotated polyp labels from over 300 patients across six independent centers. It includes both single-frame and sequence data, with precise polyp boundary delineation, aimed at advancing colon polyp detection and pixel-level segmentation research.
\item LVOS\cite{hong2023lvos} is a long-term video object segmentation dataset designed for real-world scenes, featuring videos averaging over 60 seconds. It includes 720 videos in v1 and 420 in v2\cite{hong2024lvos}, with 44 categories, providing challenges like object reappearance. The dataset is split into training, validation, and test videos to evaluate model generalization.
\item Segment Anything Video (SA-V) \cite{SAM2} is a large-scale dataset with 50,900 clips and 642,600 masks for prompt-based video segmentation. It covers challenges like small objects and occlusion, with data divided into training, validation, and test sets. This dataset supports robust model development and provides a platform for evaluating segmentation algorithms.
\end{itemize}

\subsection{Evaluate Metrics}

Video segmentation, which involves temporal data, introduces additional complexity compared to image segmentation, particularly in the temporal dimension, placing higher demands on the stability, consistency, and continuity of segmentation algorithms. While image segmentation typically focuses on pixel classification within a single frame, video segmentation must ensure both spatial segmentation and temporal coherence, avoiding issues like object jitter or drift across frames. Therefore, when evaluating video segmentation performance, it is essential not only to use common metrics such as IoU and Dice but also to consider the impact of the temporal dimension to comprehensively assess both the spatial and temporal aspects of the segmentation results.

The following metrics described in \cite{DAVIS16,DAVIS2017} are commonly used evaluation metrics for video segmentation.

\subsubsection{Region Similarity $\mathcal{J}$ and Contour Precision $\mathcal{F}$}

Given the predicted segmentation masks $\hat{M} \in \{0, 1\}^{H \times W}$ and the ground truth masks $M \in \{0, 1\}^{H \times W}$, the \textbf{region similarity} $\mathcal{J}$ is computed as the Intersection-over-Union (IoU) between $\hat{M}$ and $M$:

\begin{equation}
    \mathcal{J} = \frac{\hat{M} \cap M}{\hat{M} \cup M}.
\end{equation}

To assess the contour quality of $\hat{M}$, we calculate the contour recall $R_c$ and precision $P_c$ using bipartite graph matching \cite{martin2004learning}. The \textbf{contour accuracy} $\mathcal{F}$ is then defined as the harmonic mean of contour recall $R_c$ and precision $P_c$:
\begin{equation}
    \mathcal{F} = \frac{2 P_c R_c}{P_c + R_c}.
\end{equation}

This metric quantifies how closely the contours of the predicted masks align with those of the ground-truth masks. The average contour accuracy, $\mathcal{F}_{\text{mean}}$, is computed across all objects. For brevity, we denote this by $\mathcal{F}$.

Finally, the overall performance is measured using the \textbf{combined metric} $\mathcal{J} \& \mathcal{F}$, which is the arithmetic mean of the region similarity and contour accuracy:
\begin{equation}
    \mathcal{J} \& \mathcal{F} = \frac{\mathcal{J} + \mathcal{F}}{2}.
\end{equation}

\subsubsection{Global Accuracy}

\textbf{The $\mathcal{G}$ metric (Global Accuracy) }is used to measure the overall accuracy of segmentation results. It is defined as the proportion of correctly classified pixels in the predicted segmentation to the total number of pixels. The formula is:
\begin{equation}
    \mathcal{G} = \frac{\text{Number of correctly classified pixels}}{\text{Total number of pixels}}.
\end{equation}

Intuitively, $\mathcal{G}$ reflects the overall proportion of correctly classified pixels. A higher $\mathcal{G}$ value indicates better global segmentation accuracy.

\subsubsection{Temporal Metrics: FPS}

FPS (Frames Per Second) is a temporal metric(add cite) used to evaluate the processing speed of a video segmentation or analysis model. It measures the number of video frames that the model can process per second, reflecting its efficiency and suitability for real-time applications. A higher FPS indicates that the model can handle video data more quickly, making it ideal for time-sensitive tasks.
\begin{equation}
    \mathcal{FPS} = \frac{\text{Number of Frames Processed}}{\text{Time Taken (in seconds)}}.
\end{equation}

 \section{Discussion} \label{discussion}

During the development of this paper, we observed that SAM2 shows significant improvements compared to previous models, demonstrating great potential. However, despite these advancements, there are still several challenges and limitations that require further research and refinement. In this section, we will discuss the current challenges and opportunities in this field.

\subsection{Current Challenges}

\subsubsection{Domain Adaptation Limitations}

Although SAM2 performs well in zero-shot tasks \cite{polypSAM2,novel-sam2knee-zeroshot}, its application in domains such as medical imaging and remote sensing still requires domain-specific fine-tuning to achieve optimal performance. These fields often rely on complex contextual information~\cite{zhao2024inspiring}, and the model's generalization ability is limited without targeted data. The fine-tuning process faces challenges such as high computational costs and insufficiently labeled data, particularly for specialized tasks. These issues highlight the ongoing need for efficient domain adaptation techniques and the generation of high-quality labeled datasets.

\subsubsection{Multimodal Integration}

Another significant challenge lies in the efficient integration of SAM2 with multimodal models. SAM2 has the potential to process multimodal data, such as combining image features with textual descriptions, but the effective fusion of these data types remains a complex task \cite{samwise,huo2024sam}. Multimodal integration requires sophisticated mechanisms to align and merge data from different sources, which can be computationally intensive and may involve dealing with differences in modality-specific feature spaces. Furthermore, the model needs to effectively leverage information from diverse inputs while maintaining performance across all modalities. Future research must focus on improving multimodal interaction capabilities to ensure that SAM2 can process and understand complex multimodal data streams in a cohesive manner.

\subsubsection{Inference Speed and Resource Requirements}

SAM2, being a large and complex model, faces significant challenges~\cite{detsam2,zhang2023survey} when it comes to real-time applications, such as video segmentation and online segmentation systems. Due to its large size, SAM2 can experience slower inference times and higher resource consumption, which can hinder its deployment in environments where speed and efficiency are critical. In particular, tasks such as video segmentation, which require processing multiple frames per second, demand a high degree of computational power and low latency. Addressing these challenges will require optimization techniques that reduce the model's computational overhead while maintaining performance. Efficient resource management will be key to enabling SAM2's practical use in real-time applications, such as autonomous driving and live video analysis.

\subsection{Future Works}

Future work will focus on optimizing model performance, enhancing multimodal interaction capabilities, and improving robustness to address challenges in real-world applications.

\subsubsection{Fine-Tuning for Specialized Field}

Develop more efficient fine-tuning strategies~\cite{sam2OCTA} tailored to specific domains (e.g., medical imaging, remote sensing) to enhance the model's adaptability and performance. By leveraging domain-specific data and task-optimized techniques, the model can better address real-world applications, achieving higher precision in segmentation tasks.

\subsubsection{Lightweight Optimization}

Reduce the computational overhead of the model through techniques such as model pruning and knowledge distillation, thereby improving inference efficiency. Further optimize the model structure to ensure high performance even in resource-constrained scenarios, particularly for real-time applications~\cite{surgicalSam2,yu2024sam,lu2024samedge}.

\subsubsection{Enhanced Multimodal Interaction}

Investigate deeper integration of SAM2 with multimodal data inputs, such as language descriptions and textual information~\cite{zhang2024evf,zhu2024customize,fusionsam}, to explore its potential application scenarios. By enhancing multimodal interaction, the model can be applied more effectively to complex and diverse tasks, including intelligent question answering and image-text analysis.

\subsubsection{Improving Robustness}

Incorporate a wider variety of complex and diverse datasets during training to improve the model's ability to handle challenging scenarios such as noise and occlusion~\cite{shen2024performance,rafaeli2024prompt}. Techniques such as data augmentation and adversarial training can further enhance the model’s stability and reliability under uncertain conditions. 
 \section{Conclusion} \label{conclusion}

This paper reviews the advances and challenges of SAM2 in the field of image and video segmentation. Compared to its predecessor, SAM2 shows significant improvements in handling complex scenarios, though its performance in specific domains such as medical imaging and autonomous driving still requires further optimization. The study focuses on SAM2's applications in image and video segmentation: in the image domain, it emphasizes its capabilities in medical image processing; in the video domain, it highlights its handling of temporal consistency. Despite existing challenges, the technological potential of SAM2 offers valuable directions for future research and practical applications. We hope that the insights provided in this paper will serve as a useful reference for researchers, driving the continued optimization and broader adoption of SAM2 in the field of computer vision.

\small
\bibliographystyle{IEEEtran}
\bibliography{ref}

\end{document}